\documentclass{article}

\usepackage{hyperref}
\usepackage[accepted]{icml2025}
\usepackage[textsize=tiny]{todonotes}

\usepackage[utf8]{inputenc}
\usepackage[T1]{fontenc}    
\usepackage{hyperref}       
\usepackage{url}            
\usepackage{booktabs}       
\usepackage{amsfonts}       
\usepackage{nicefrac}       
\usepackage{microtype}      
\usepackage{xcolor}         
\usepackage{soul}

\usepackage{graphicx}
\usepackage{amsmath}
\usepackage{mathtools}
\usepackage{bm}
\usepackage{bbm}
\usepackage{subcaption}
\graphicspath{ {./Figures/} }
\usepackage{cancel}
\usepackage{enumitem}
\newlist{stepenumerate}{enumerate}{1}
\setlist[stepenumerate,1]{label=\arabic*. , start=0}

\usepackage{xspace}
\newcommand{\acro}[1]{\textsc{#1}\xspace}
\newcommand{\cotp}{{\textsc {CoT}}\xspace}

\newcommand{\cladder}{{\textsc {CLadder}}\xspace}
\newcommand{\ctoc}{{\textsc{Corr2Cause}}\xspace}
\newcommand{\ourprompting}{{\textsc{PC-SubQ}}\xspace}
\newcommand{\llm}{\acro{LLM}}
\newcommand{\cd}{\acro{CD}}
\newcommand{\scd}{\acro{S-CD}}
\newcommand{\pc}{\acro{PC}}
\newcommand{\dagr}{\acro{DAG}}
\newcommand{\mec}{\acro{MEC}}
\newcommand{\cpdag}{\acro{CPDAG}}
\newcommand{\nlcd}{\acro{NL-CD}}
\newcommand{\subq}{{\textsc {SubQ}}\xspace}
\DeclareBoldMathCommand{\X}{X}
\DeclareBoldMathCommand{\E}{E}
\DeclareBoldMathCommand{\Z}{Z}
\DeclareBoldMathCommand{\Y}{Y}
\DeclareBoldMathCommand{\A}{A}
\DeclareBoldMathCommand{\B}{B}
\newcommand{\ind}{\perp\!\!\!\!\perp}

\newcommand{\cond}{\,|\,}

\usepackage[super]{nth}

\usepackage[capitalize,noabbrev]{cleveref}
\icmltitlerunning{Prompting Strategies for LLMs to Infer Causation from Correlation}

\begin{document}

\twocolumn[
\icmltitle{Prompting Strategies for Enabling Large Language Models\\ to Infer Causation from Correlation}

\begin{icmlauthorlist}
\icmlauthor{Eleni Sgouritsa}{GDM}
\icmlauthor{Virginia Aglietti}{GDM}
\icmlauthor{Yee Whye Teh}{GDM}
\icmlauthor{Arnaud Doucet}{GDM}
\icmlauthor{Arthur Gretton}{GDM}
\icmlauthor{Silvia Chiappa}{GDM}
\end{icmlauthorlist}

\icmlaffiliation{GDM}{Google DeepMind, London, UK}

\icmlcorrespondingauthor{Eleni Sgouritsa}{eleni.sgouritsa@google.com}

\icmlkeywords{causality, prompting strategies} 

\vskip 0.3in
]

\printAffiliationsAndNotice{}

\begin{abstract}
The reasoning abilities of Large Language Models (\llm{s}) are attracting increasing attention. In this work, we focus on causal reasoning and address the task of establishing causal relationships based on correlation information, a highly challenging problem on which several \llm{s} have shown poor performance. We introduce a prompting strategy for this problem that breaks the original task into fixed subquestions, with each subquestion corresponding to one step of a formal causal discovery algorithm, the \pc algorithm. The proposed prompting strategy, \ourprompting, guides the \llm to follow these algorithmic steps, by sequentially prompting it with one subquestion at a time, augmenting the next subquestion's prompt with the answer to the previous one(s). 
We evaluate our approach on an existing causal benchmark, \ctoc: our experiments indicate a performance improvement across five \llm{s} when comparing \ourprompting to baseline prompting strategies. Results are robust to causal query perturbations, when modifying the variable names or paraphrasing the expressions.
\end{abstract}

\section{Introduction}

Advances in scaling Large Language Models (\llm{s}) and training data sizes have led to unprecedented capabilities across a wide range of tasks.
Several works \citep{huang2022towards, sun2023survey} focus on enhancing \llm{s} on various reasoning tasks, such as arithmetic, commonsense, logical or causal reasoning.
It is often argued that the ability to demonstrate reasoning cannot be solely overcome by increasing model scale \citep[\textit{inter alia}]{rae2021scaling}. 
An active area of research is therefore investigating prompting approaches in order to elicit and enhance reasoning in \llm{s} \citep{qiao2023reasoning}. \citet{wei2022chain} proposed Chain-of-Thought (\cotp) prompting, where a language model is prompted to generate a series of thoughts that mimic the reasoning process a person might employ when solving a task. For settings in which the problem to solve is harder than the demonstration examples, various works suggest to decompose the original complex task into simpler subtasks that are easier to solve \citep{zhou2023least, khot2023decomposed, dua2022successive}.

Among the different reasoning capabilities, the ability to reason about cause and effect is fundamental to human intelligence, playing a pivotal role in scientific understanding and accurate decision-making. 
Most existing research on causal reasoning and \llm{s} investigates whether they can serve as a knowledge base for causal relationships \citep{kiciman2023causal, willig2022can, zevcevic2023causal, long2022can}, 
thus assessing the \textit{commonsense reasoning} capabilities of \llm{s}. 
More recent work \citep{jin2023cladder,jin2024can} focuses, instead, on their \textit{formal (pure) causal reasoning} skills, i.e., their ability to reason using known procedures and formal rules from the causal inference literature \citep{pearl2000causality, pearl2018book, spirtes2000causation}. One such ability, studied in \citet{jin2024can}, is that of \textit{inferring causal relationships from correlation statements}. In this case, 
correlations or conditional independencies about a set of variables are given in natural language, for instance, ``Ice cream sales correlate with shark attacks'', and the validity of a causal relationship between a pair of variables is inferred, for instance, whether ``Ice cream sales cause shark attacks''. This is an important skill for \llm{s} as the training corpus may not directly contain certain causal relationships between variables, but instead events' co-occurrence or correlation statements. To this end, \citet{jin2024can} formulated {\ctoc}, a causal benchmark that assesses this ability, on which they showed \llm{s} to perform poorly.

Expanding the task of inferring causal relationships from correlation statements, one could more generally \textit{infer the causal structure from correlation statements}, that is, identify the entire 
structure instead of a specific causal relationship between a pair of variables. 
We refer to this task as \textit{Natural Language Causal Discovery} (\nlcd) 
to contrast it with standard/statistical \textit{Causal Discovery} (\scd) considered in the statistical causality literature \citep{pearl2000causality, pearl2018book, spirtes2000causation}, that identifies the causal graph based on observed data. \nlcd, instead, is an \llm task that is based on natural language correlation statements. 
Even though there exist known statistical algorithms for certain causal tasks, our goal is to instead investigate the ability of \llm{s} to solve such tasks. This motivation is also shared in \citet{jin2024can} and \citet{jin2023cladder} for causal tasks, in \citet{markeeva2024clrs} and \citet{velivckovic2022clrs} for algorithmic tasks and in other works for arithmetic tasks.
In this work, we address \nlcd by introducing a prompting strategy, \ourprompting.
Inspired by prompting approaches that decompose a complex task into simpler ones, \ourprompting leverages one of the most widely used algorithms for \scd, the \pc algorithm \citep{spirtes2000causation}, in order to decompose the \nlcd task. 
In particular, \ourprompting decomposes the original task into several \textit{fixed} subtasks, each corresponding to one step of the \pc algorithm. We guide the \llm to follow these \pc steps by having few-shot examples for each subtask and sequentially prompting the \llm, augmenting the next subtask with the answer(s) to the previous one(s). Given the final inferred graph, one can subsequently pose various causal queries, such as the ones included in \ctoc.
We assess \ourprompting on \ctoc and consistently observe improved performance compared to existing prompting strategies across five \llm{s}. 
Moreover, unlike models fine-tuned on \ctoc \citep{jin2024can}, \llm{s'} performance using \ourprompting is robust to query perturbations: even though the few-shot examples are based on \ctoc textual expressions, performance does not drop when variable names are modified or expressions are paraphrased. 
In summary, our main contributions are:
\vspace{-0.3cm}
\begin{itemize}
    \item We propose a prompting strategy that addresses \nlcd by guiding the \llm to reason through applying the \pc algorithm steps, thus also offering transparent and interpretable reasoning steps.
    \vspace{-0.245cm}
    \item We evaluate \ourprompting  on \ctoc and show that it outperforms a range of widely used prompting strategies across five \llm{s}.
    \vspace{-0.245cm}
    \item We show that \ourprompting is robust to causal query perturbations in terms of variable renaming and paraphrasing, and can be applied unchanged to examples with natural stories.
\end{itemize}

\section{Related Work} \label{sec:related}

\paragraph{Causal reasoning and \llm{s}} 
There is an increasing attention on the strengths and limitations of \llm{s} on causal tasks \citep{kiciman2023causal, willig2022can, zevcevic2023causal, zhang2023understanding}.
A large body of existing literature assesses how well \llm{s} capture commonsense and domain knowledge about causal relations \citep{kiciman2023causal, long2022can, zhang2023understanding, tu2023causal, vashishtha2023causal, jiralerspong2024efficient}.
Consider, for example, the BIG-bench Cause and Effect task of deciding which of the two events, "The driver turned the wipers on" and "It started raining", caused the other \citep{big-bench2023beyond}. This question could be answered by using commonsense knowledge about the purpose of wipers.
Commonsense knowledge captured by \llm{s} can be used alongside existing causal methods as a proxy for human domain knowledge \citep{kiciman2023causal, long2022can, long2023causal}. Indeed, recent studies suggest to improve \scd methods either by incorporating \llm{s'} imperfect knowledge about causal relationships \citep{long2023causal, abdulaal2024causal} or through providing an \llm-based oracle of conditional independence relations as input to the \pc algorithm \citep{cohrs2023large}.
On a more critical perspective, \citet{willig2022can} and \citet{zevcevic2023causal} argue that \llm{s} are just ``causal parrots'' which recite the causal knowledge embedded in the training data, and so we cannot expect any sort of generalisation. 

Another way of inferring causal relationships is through formal causal reasoning, that is, by employing known rules and algorithms from the causal inference literature \citep{pearl2000causality,spirtes2000causation} to reach a causal conclusion. In this case, the inference rules hold independently of the specific instantiation of the variables, as opposed to commonsense reasoning which highly relies on the semantics of the variable names.
\citet{jin2024can} and \citet{jin2023cladder} analyse the difference between these two reasoning abilities, i.e., commonsense and formal causal reasoning, and introduce benchmarks, \ctoc and \cladder, respectively, to test the formal causal reasoning abilities of \llm{s}. Specifically, \ctoc consists of queries about causal relationships given correlation statements, while \cladder asks causal inference queries across all rungs of the Ladder of Causation \citep{pearl2018book} for a given underlying causal graph. Both works observe that it is highly challenging for a range of \llm{s} to answer these causal queries. 
In particular, \citet{jin2024can} mainly evaluate off-the-shelf or fine-tuned \llm{s} on \ctoc and find them to perform poorly or not to be robust to causal query perturbations. 
We instead propose a multi-step prompting strategy (\ourprompting) that is robust to causal query perturbations, produces interpretable reasoning steps and does  not require \llm retraining.

\vspace{-0.15cm}
\paragraph{Prompting strategies} 
Since the popularization of few-shot prompting \citep{brown2020language}, there has been a lot of follow-up work on prompting strategies. In Chain-of-Thought (\cotp) prompting \citep{wei2022chain}, a few chain of thought demonstrations are provided as examples in the prompt. This significantly improves the ability of \llm{s} to perform complex reasoning on a range of tasks, with examples of surpassing even fine-tuned models. 
To address more complicated tasks, several follow-up works suggest to decompose the original complex task into simpler sub-tasks that are easier to solve \citep{zhou2023least,khot2023decomposed,dua2022successive}.
Least-to-most prompting \citep{zhou2023least} solves these simpler problems in sequence, with each subproblem being facilitated by previously solved ones. 
Unlike least-to-most prompting that infers the subquestions using an appropriate prompt, we instead design a \textit{fixed} set of subquestions that is used unchanged across the whole dataset.
For that, we take advantage of the fact that there is a common underlying (causal) algorithm that can be followed to solve all causal queries, so we design the subproblems accordingly to correspond to the \pc steps. 
\citet{jin2023cladder} use a prompting strategy comprising of a chain of subquestions without demonstrations, in order to answer various causal query types based on a given causal graph.
Unlike the above works, we do not carry over to the next subquestion all history of previous ones, but propagate much less information (usually just the final answer to the previous subquestion) in order to save context length and guide the \llm to focus on the most relevant parts of information. Finally, our work connects to literature on algorithmic reasoning \citep{zhou2023teaching, sel2023algorithm, nye2021show}. \citet{zhou2023teaching} use an algorithmic prompt to unlock algorithmic reasoning abilities in \llm{s}.

\section{Preliminaries} \label{sec:preliminaries}

\paragraph{Graph definitions}
A graph ${\cal G} \coloneqq \langle \X,\E\rangle $ consists of a set of nodes 
$\X$ and a set of edges $\E$ connecting the nodes. An \textit{undirected graph} consists of undirected edges denoted as $X - Y$, while a \textit{directed graph} has directed edges denoted as $X \rightarrow Y$. A \textit{path} from $X$ to $Y$ is a sequence of distinct nodes $(X,\ldots,Y)$ in which there is an edge between every pair of successive nodes. A \textit{directed path} is a path whose edges are directed and pointing from preceding towards following nodes. A \textit{Directed Acyclic Graph} (\dagr) is a directed graph with no cycles, i.e. no directed paths starting and ending at the same node. The \textit{skeleton} of a \dagr is the undirected graph resulting from ignoring all arrowheads.
When $X \rightarrow Y$, $X$ is called the \textit{parent} of $Y$ and, consequently, $Y$ the \textit{child} of $X$. A \textit{collider} is a node where at least two arrowheads meet, for example, in the path $(X, Z, Y)$, $Z$ is a collider if $X \rightarrow Z$ and $Y \rightarrow Z$. A \textit{v-structure} consists of two edges whose arrows point to the same node and whose tails are not connected by an edge: for example, the path $(X, Z, Y)$ forms a v-structure if $X \rightarrow Z$, $Y \rightarrow Z$ and $X$ and $Y$ are not connected. 
A path between two nodes is said to be \textit{unblocked} conditioned on a set of nodes $\Z$, if for every collider $W$ in the path, either $W$ or a descendant of $W$ is in $\Z$ and no non-collider in the path is in $\Z$. A \textit{blocked} path is a path that is not unblocked. Two disjoint sets of nodes $\A$ and $\B$ are said to be \textit{d-separated} given a set of node $\Z$ if every path between any node in $\A$ and any node in $\B$ is blocked conditioned on $\Z$.

\vspace{-0.15cm}
\paragraph{Markov condition} Consider the nodes in a \dagr $\cal{G}$ representing random variables with $P(\X)$ denoting their joint distribution.\footnote{For simplicity, we henceforth refer to a \dagr with its nodes representing random variables jointly distributed according to $P(\X)$ simply as a \dagr.} The joint distribution is \textit{Markov} to $\cal{G}$ if the following condition, called the \textit{Markov condition}, holds for all $\A, \B, \Z$:
$\A$, $\B$ d-separated given $\Z$ in $\cal{G}$ $\Rightarrow$ $\A \ind \B \cond \Z$.
In other words, the Markov condition entails a set of conditional independence relations which can be found by the d-separations encoded in the graph. Two \dagr{s} are called Markov equivalent or alternatively said to belong to the same Markov Equivalence Class (\mec) if they entail the same set of conditional independence relations and can be compactly represented as a partially directed graph, so-called Completed Partially Directed Acyclic Graph (\cpdag).

\vspace{-0.15cm}
\paragraph{Causal discovery}
A causal \dagr is a \dagr in which edges represent causal influence, with $X \rightarrow Y$ denoting that $X$ is a direct cause of $Y$.
Causal Discovery (\cd) is the task of (possibly partially) learning the causal \dagr between a set of variables based on a data sample from their joint distribution. \cd methods can be categorised into constraint-based \citep{spirtes2000causation}, score-based-methods \citep{chickering2002learning} and methods restricting the function class \citep{shimizu2006linear, hoyer2008nonlinear, peters2014causal}. Constraint-based methods rely on conditional independence tests between the variables $\X$ in order to rule-out edges in the constructed \dagr, assuming faithfulness, that is that the only conditional independencies are those implied by d-separation \citep{spirtes2000causation}. The Peter-Clark (\pc) algorithm \citep{spirtes2000causation} is one of the most widely used constraint-based methods, that assumes no unobserved confounders (direct common causes of two or more measured variables).
Its final output is a \cpdag or equivalently a \mec, including \dagr{s}, entailing the same set of conditional independence relations, which cannot be further distinguished without additional assumptions.

\vspace{-0.15cm}
\paragraph{\ctoc benchmark}
We provide a description of the \ctoc benchmark \citep{jin2024can}, which we use in our evaluations.
Each \ctoc record consists of three parts: a \textit{Premise} consisting of correlation statements, a \textit{Hypothesis} with the causal question and a binary \textit{Output Label} to be predicted indicating whether the hypothesis is valid ($1$) or invalid ($0$). For example:
\vspace{-0.3cm}
\begingroup
\addtolength\leftmargini{-0.07in}
\begin{quote}
\textit{Premise}: Suppose there is a closed system of 3 variables $A$, $B$ and $C$. All the statistical relations among these 3 variables are as follows: $A$ correlates with $C$. $B$ correlates with $C$. However, $A$ is independent of $B$. \textit{Hypothesis}: $A$ directly causes $C$.\\
\textit{Output Label}: $1$
\end{quote}
\endgroup
\vspace{-0.2cm}

\begin{figure*}[ht!]
    \centering
      \includegraphics[width=0.91\textwidth]{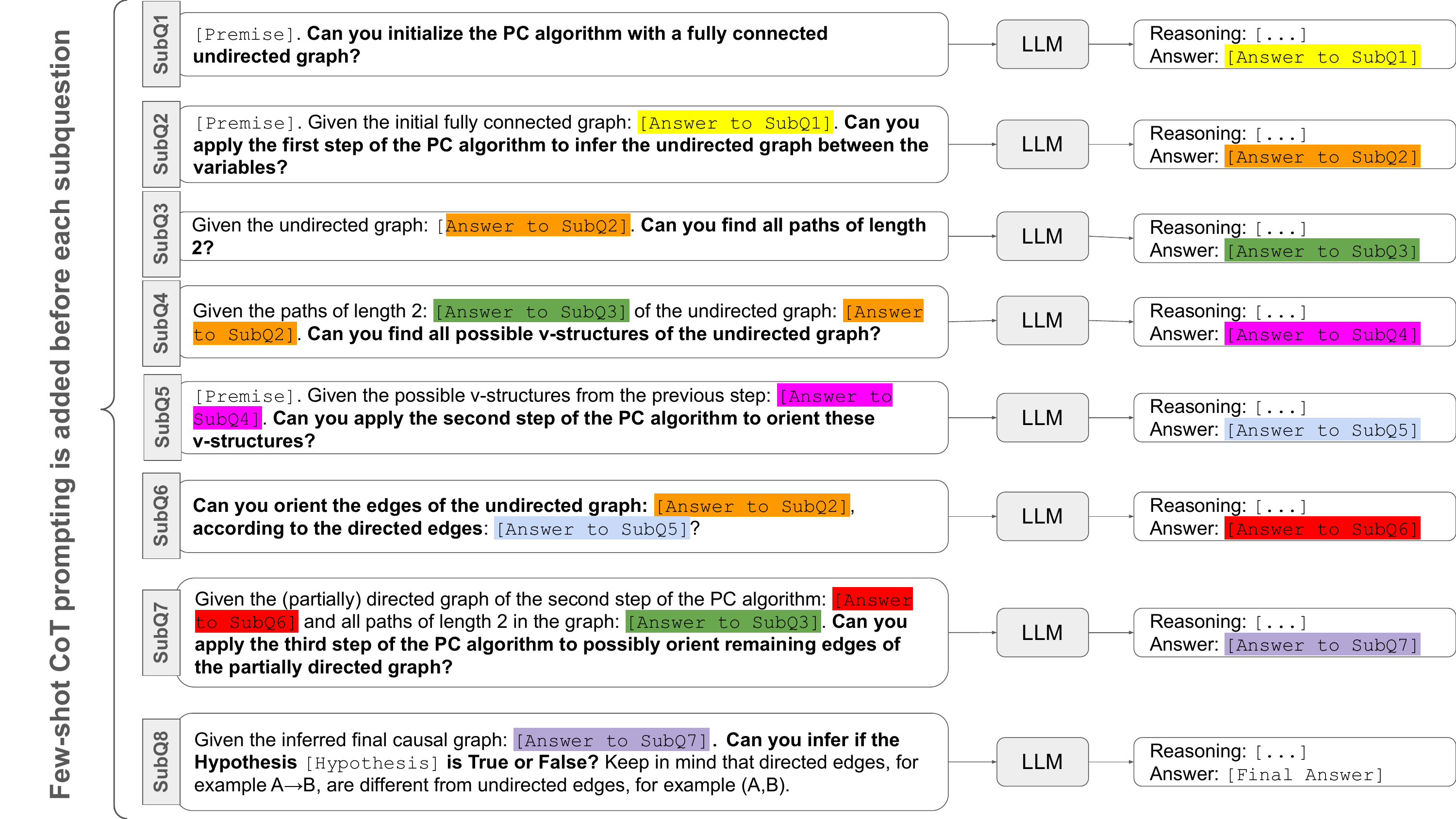}
      \caption{The 8 fixed subquestions of \ourprompting.
      \texttt{[Premise]} and \texttt{[Hypothesis]} are placeholders for the input Premise and Hypothesis, respectively, while \texttt{[Answer to SubQi]} and \texttt{[Final Answer]} are placeholders for the output intermediate and final answers, respectively. \texttt{[...]} represents some reasoning text that we expect from the \llm. The colors demonstrate how the answers to previous subquestions are passed as input to the next ones. Few-shot \cotp examples are prepended to each subquestion (see Fig. \ref{fig:subquestions_shots})} \label{fig:subquestions}
\end{figure*}

The benchmark is constructed as follows. First, all unique \dagr{s} with $N=2,\ldots,6$ nodes are generated. For every \dagr, the Markov condition implies certain conditional independencies based on which \dagr{s} are clustered into \mec{s}. For each \mec, several causal hypotheses between every pair of variables are considered, such as direct causation (as in the \textit{Hypothesis} of the example above), confounding or mediating.
Given a set of conditional independencies and a causal hypothesis as input, the output label represents the validity of the hypothesis and is generated as follows: if all \dagr{s} in the \mec (implying the given conditional independencies) satisfy the given hypothesis, then the hypothesis is deemed valid, otherwise it is deemed invalid.
Everything is, then, verbalised into natural language, with the verbalised sets of conditional independencies comprising the \textit{Premise}. Variable names are kept symbolic, i.e., $A, B, C, \ldots$, as in the example above, to disentangle the formal from the commonsense reasoning abilities of \llm{s}.
For a detailed description on the dataset construction one could refer to \citet{jin2024can}.

\section{Proposed Prompting Strategy: \ourprompting}

\begin{figure*}[t!]
      \centering
      \includegraphics[width=0.90\textwidth]{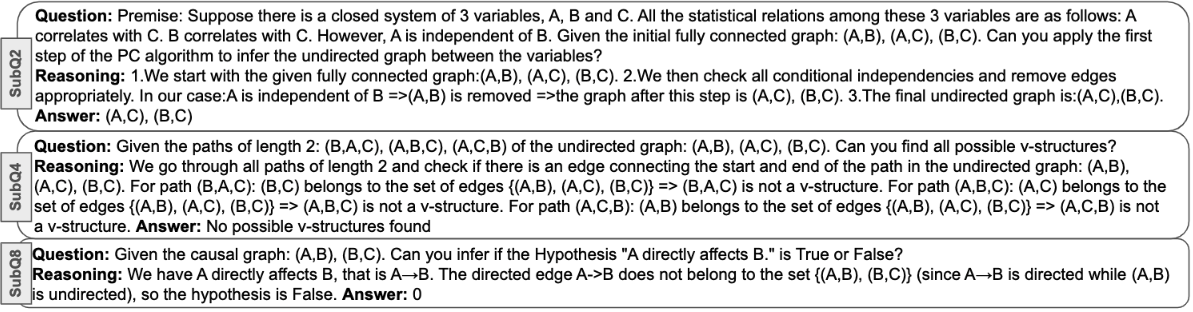}
      \caption{Indicative few-shot examples prepended to \ourprompting subquestions (see Fig. \ref{fig:subquestions}).}
    \label{fig:subquestions_shots}
\end{figure*}

We address the \nlcd task of inferring causal structure from correlation statements.
Prompting an \llm to decompose a complex problem into simpler ones was recently shown to improve performance \citep{zhou2023least,khot2023decomposed,dua2022successive}. Inspired by this, we propose a prompting strategy, \ourprompting, which exploits the steps of a popular \cd algorithm, the \pc algorithm \citep{spirtes2000causation}, in order to decompose the \nlcd task.
In particular, we break the original task into several \textit{fixed} subquestions, each corresponding to one step of the \pc algorithm (Fig. \ref{fig:subquestions}).
We explicitly guide the \llm to follow these steps, by sequentially prompting the \llm with one \pc subquestion at a time, augmenting the next subquestion's prompt with the answer(s) to the previous one(s). 
Few-shot \cotp examples \citep{wei2022chain} are provided for every subquestion with demonstrations of how to solve this specific subquestion (Fig. \ref{fig:subquestions_shots}). Subquestions and corresponding few-shots examples are used unchanged across all \nlcd queries, since \pc is a common algorithm that can be followed to solve all such causal queries.

Figure \ref{fig:subquestions} shows the 8 fixed \ourprompting subquestions that we designed.
\ourprompting sequentially prompts the \llm with one of these 8 subquestions at a time.
The coloring scheme demonstrates how answers to previous subquestions are passed to the next ones, augmenting subsequent prompts.
Notice that the rest of the history, for instance the reasoning steps, is not used in the prompt construction of following subquestions. This allows reducing the context length while enabling the \llm to focus on the most relevant information for solving the specific sub-task at hand.
\texttt{[Premise]} and \texttt{[Hypothesis]} refer to the given conditional independence or correlation statements and to the queried hypothesis about causal relationships, respectively (see also \ctoc in Sec. \ref{sec:preliminaries}).
In the following, we first provide the steps of the \pc algorithm \citep{spirtes2000causation} and then their correspondence to our subquestions:
\vspace{-0.22cm}
\begin{stepenumerate}
    \setlength\itemsep{0.08em}
    \item \textbf{Initialise fully connected graph:}
    Construct a complete undirected graph of all variables.
    \vspace{-0.1cm}
    \item \textbf{Construct skeleton:} Eliminate as many edges as possible using conditional independence relations: remove edge $X - Y$ if $X$ is independent of $Y$ given some conditioning set $Z$.
    \vspace{-0.1cm}
    \item \textbf{Orient edges to form v-structures:} For all paths $(X,Z,Y)$, such that edge $X - Y$ is absent and $Z$ is not in any conditioning set that makes $X$ and $Y$ independent: orient arrows such that a v-structure is created, i.e., $X \rightarrow Z$ and $Y \rightarrow Z$.
    \vspace{-0.1cm}
    \item \textbf{Orient edges to not form extra v-structures:} Possibly orient some more edges such that no extra v-structures are introduced: For paths $(X,Z,Y)$ with only one of the two edges oriented towards $Z$, orient the other edge away from $Z$.
\end{stepenumerate}
\vspace{-0.22cm}

The correspondence between subquestions (\subq) and \pc algorithm steps is as follows: \subq1 refers to step 0 of the \pc algorithm, \subq2 to step 1, \subq3-6 to step 2 and \subq 7 to step 3. The third step of the \pc algorithm, orienting v-structures, is the most complicated one, thus it was split into more fine-grained steps corresponding to four subquestions.
Overall, \subq 1-7 correspond to the four steps of the \pc algorithm, and aim to guide the \llm to infer the (possibly partially directed) causal graph. 
The last \subq8 questions, instead, the validity of the causal hypothesis between a pair of variables given the causal graph inferred from the previous seven subquestions.
We prepend few-shot examples, each with a \cotp reasoning, to each subquestion. We use 1-4 shots per subquestion except for \subq8 that uses 11 shots with diverse examples covering a range of possible causal hypotheses.
Figure \ref{fig:subquestions_shots} depicts some indicative shots for some of the subquestions, while all few-shot examples are provided in Appendix \ref{sec:appendix_subquestions_shots}.
Notice that the few-shot \cotp examples that are provided for one subquestion are ``independent'' of the examples provided for another, in the sense that they only demonstrate how to solve this \textit{specific} subquestion and not any of the previous ones. This allows the demonstrations for two subquestions to not necessarily refer to the same example query: for instance, in Fig. \ref{fig:subquestions_shots}, the shot for \subq4 refers to 3 variables, whereas the shot for \subq5 refers to 4 variables.

\begin{figure*}[!t]
    \centering
    \includegraphics[width=0.81\textwidth]{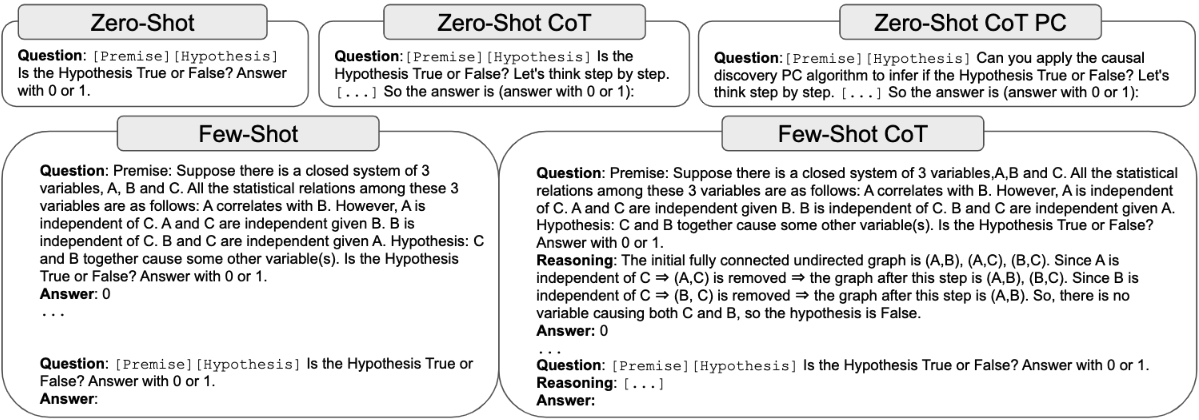}
    \caption{Five prompting strategies to compare \ourprompting with. \texttt{[Premise]} and \texttt{[Hypothesis]} are placeholders for the Premise and Hypothesis, respectively, while \texttt{[...]} represents some reasoning text that we expect from the \llm. For both few-shot strategies only one indicative example is shown.}
    \label{fig:baselines}
\end{figure*}

\begin{figure*}[t!]
    \centering
    \begin{subfigure}{0.37\textwidth}
    \includegraphics[width=\textwidth]{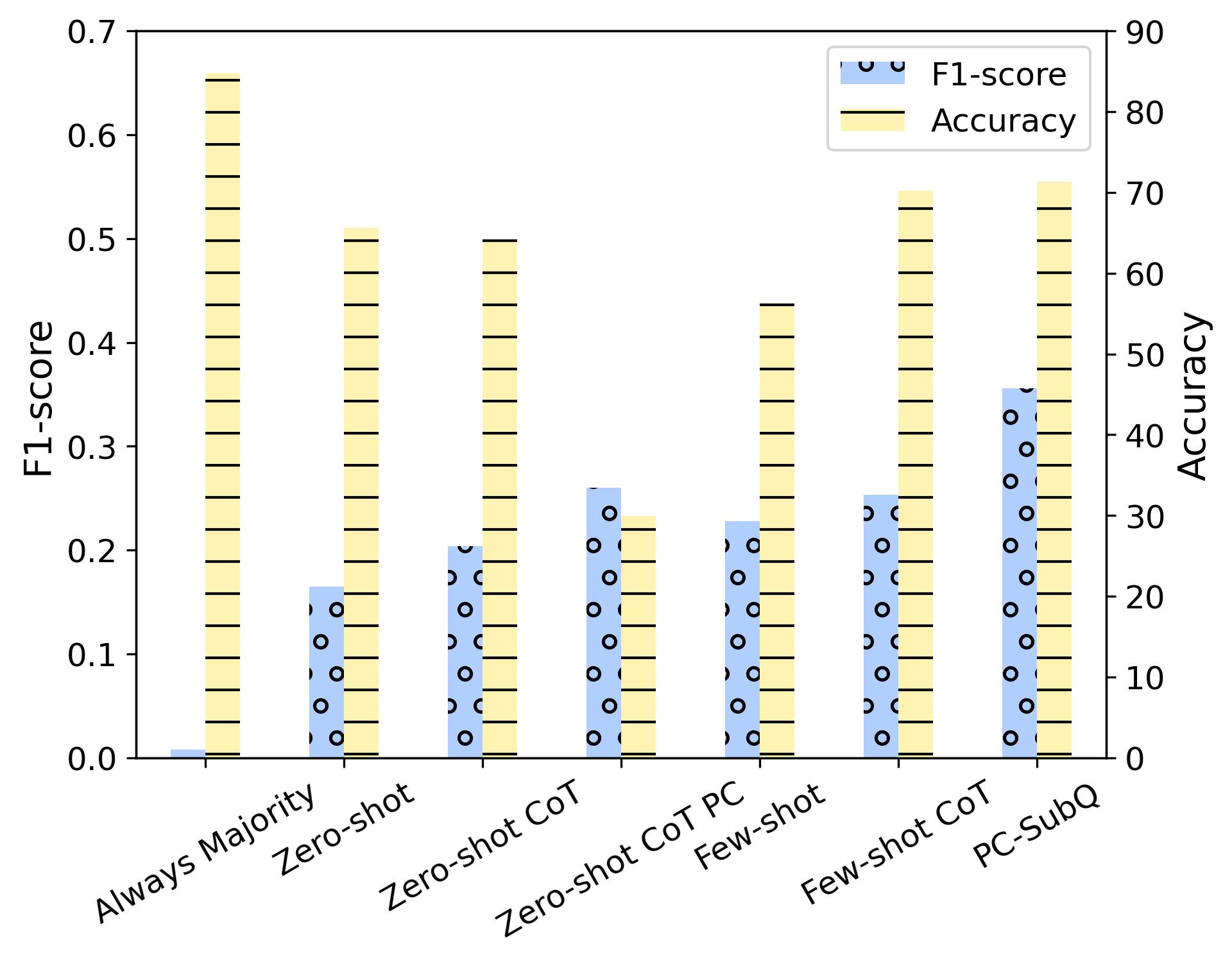}
    \caption{}
    \end{subfigure}
    \begin{subfigure}{0.37\textwidth}
    \includegraphics[width=\textwidth]{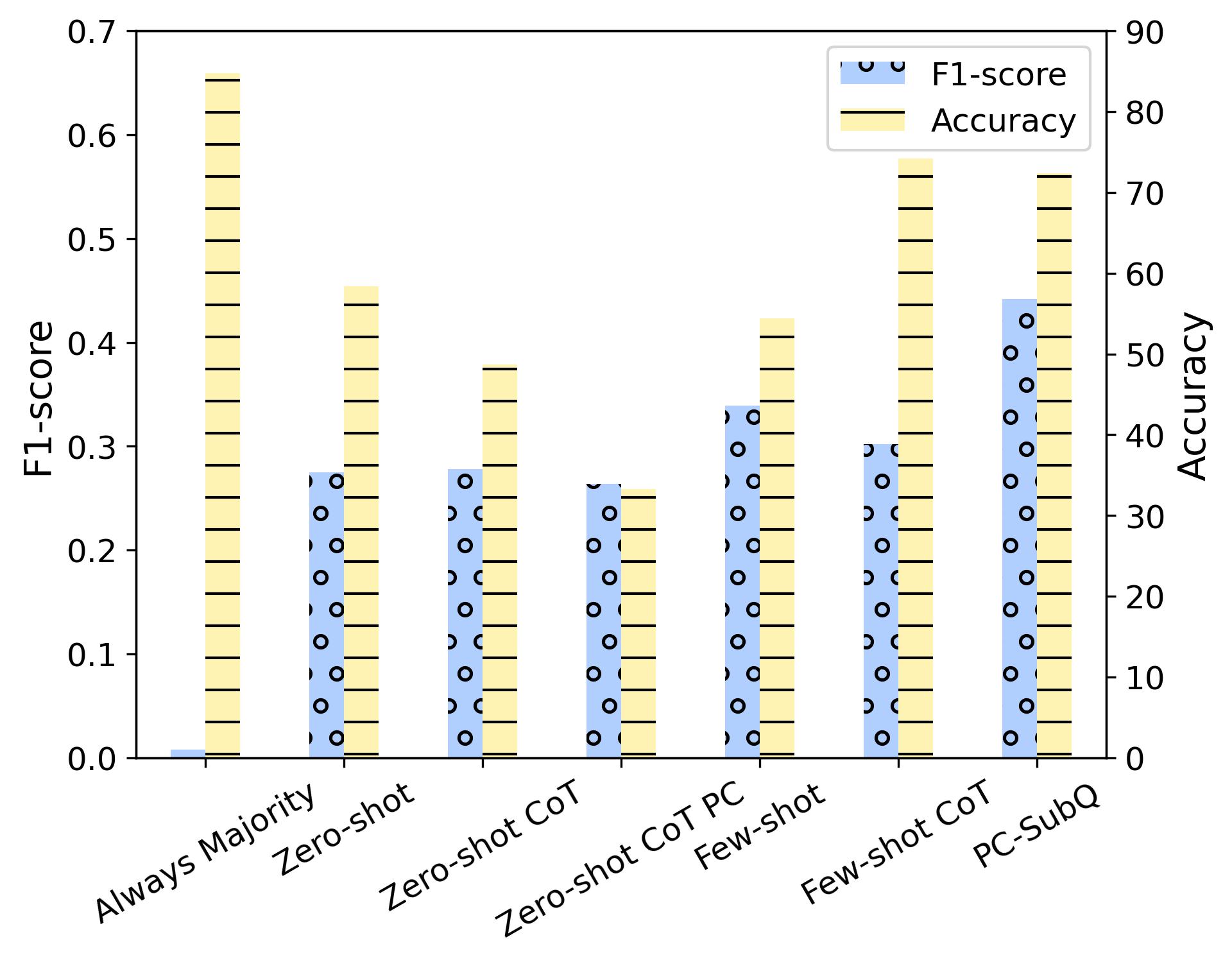}
    \caption{}
    \end{subfigure}
    \begin{subfigure}{0.27\textwidth}
    \includegraphics[width=\textwidth]{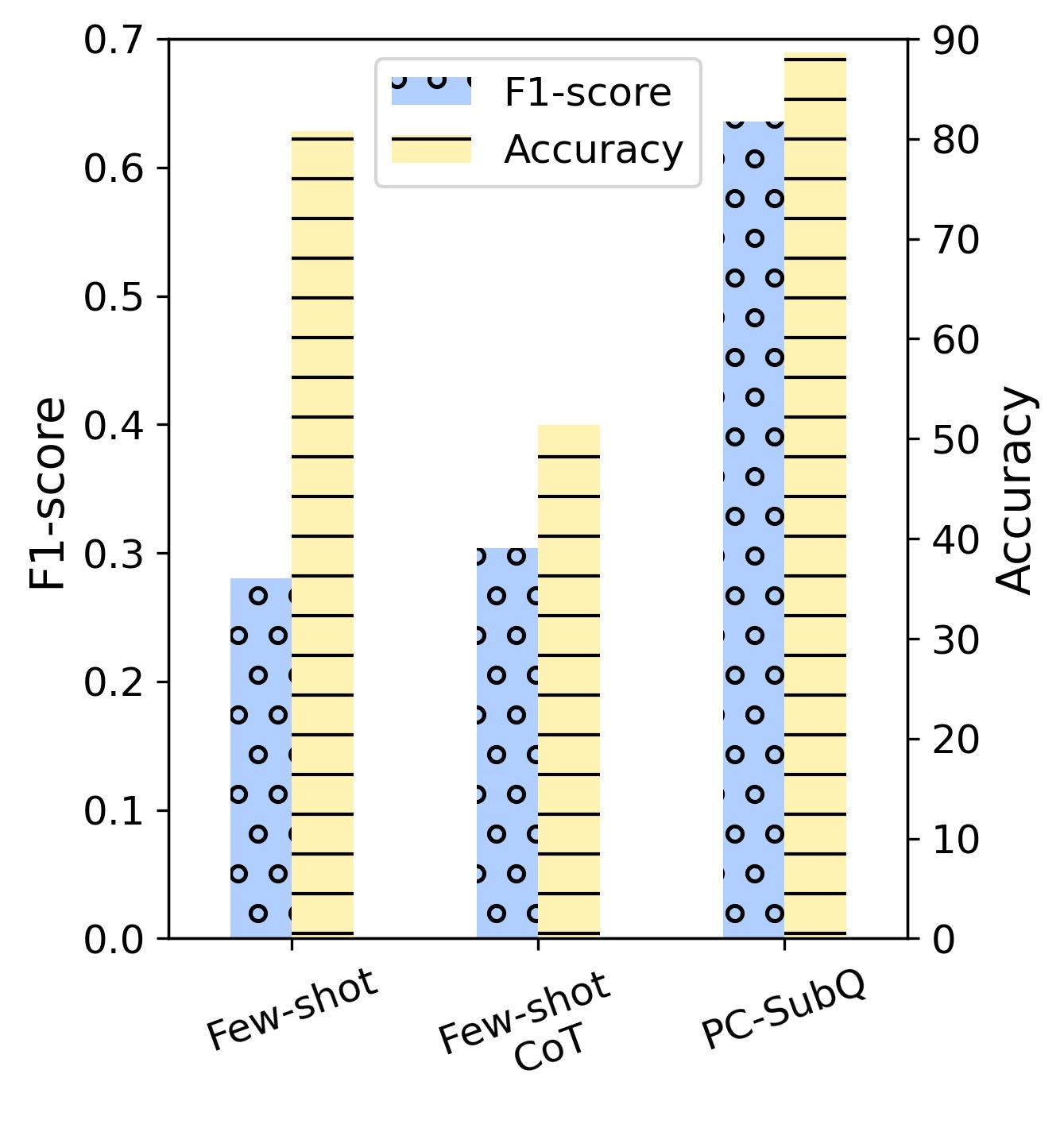}
    \caption{}
    \end{subfigure}
    \begin{subfigure}{0.27\textwidth}
    \includegraphics[width=\textwidth]{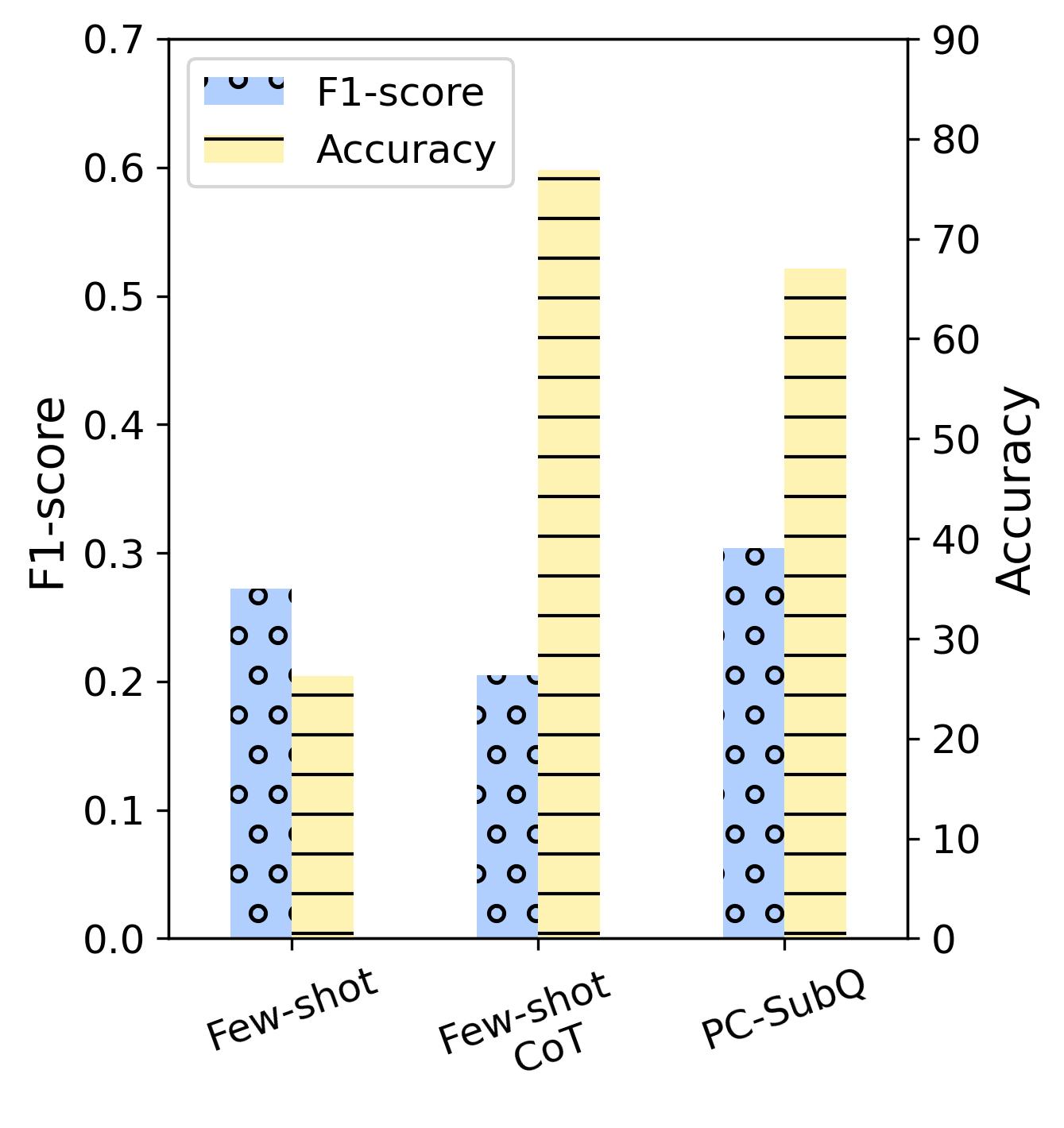}
    \caption{}
    \end{subfigure}
    \begin{subfigure}{0.27\textwidth}
    \includegraphics[width=\textwidth]{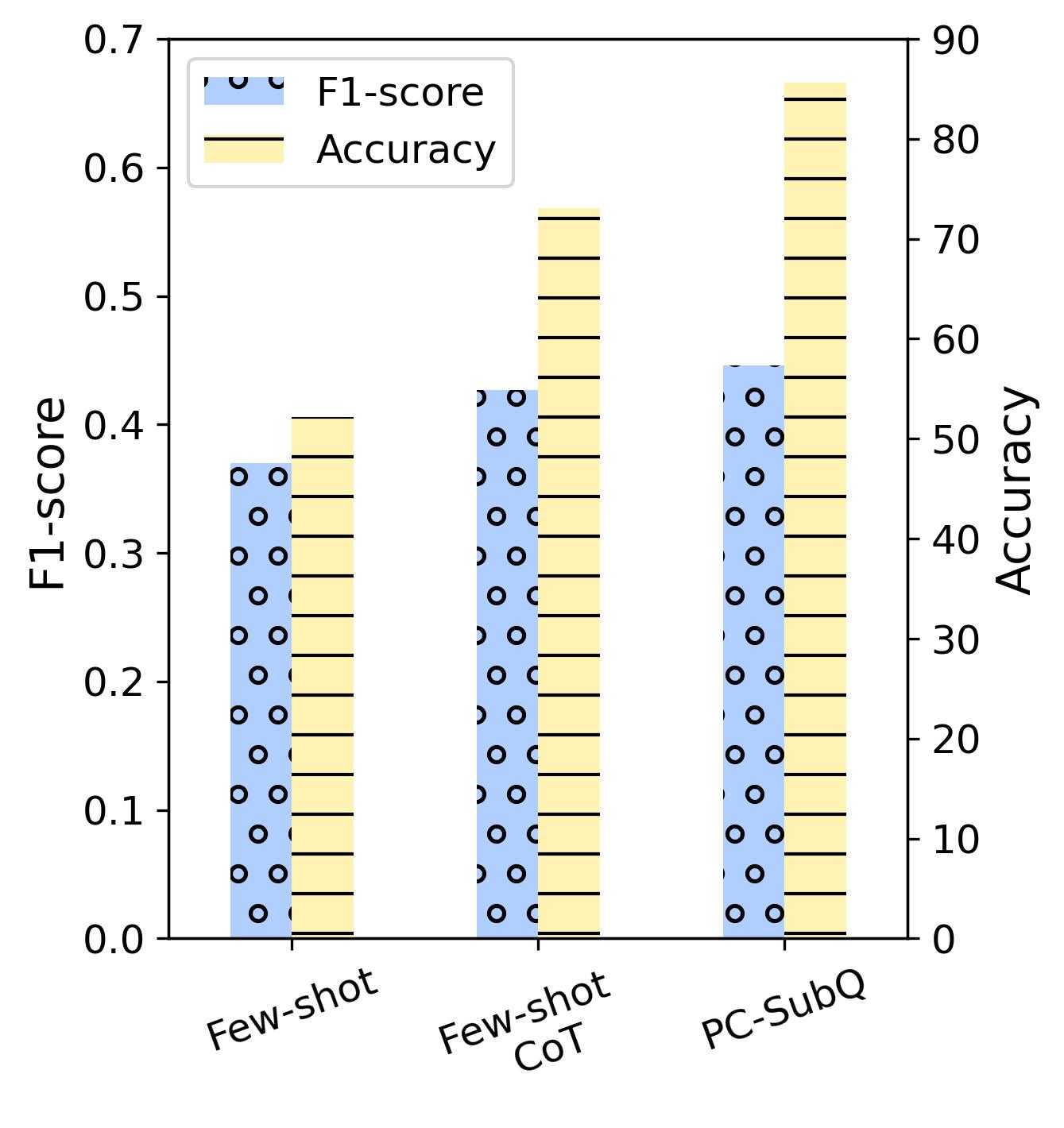}
    \caption{}
    \end{subfigure}
    \caption{F1-score and accuracy metrics for a range of prompting strategies using (a) Gemini Pro 1.0., (b) Gemini Ultra 1.0, (c) PaLM 2 L, (d) GPT-3.5-turbo and (e) GPT-4-turbo. \ourprompting outperforms all other prompting strategies as measured by our main metric (F1-score).}
    \label{fig:results_all_models}
\end{figure*}

Even though the few shots prepended to each subquestion (see Fig.\ref{fig:subquestions_shots}) are based on \ctoc,
\ourprompting can be used unchanged, i.e., with the subquestions and corresponding few shots fixed, with other benchmarks assessing \llm{s'} \nlcd skills. For instance, benchmarks containing natural variable names/scenarios, instead of symbolic, could be considered more appealing. 
A case study towards creating such a benchmark is provided in \citet{jin2024can} and, in the next section, we provide \ourprompting{'s} output on such natural causal queries. Moreover, \ourprompting is robust to perturbations of the causal queries, such as different instantiation of the variables or paraphrasing (see Sec. \ref{sec:results}).

\section{Results} \label{sec:results}

In the following, we first present results with our proposed \ourprompting strategy on \ctoc using a diverse list of five \llm{s} and compare them to the baseline prompting strategies depicted in Fig. \ref{fig:baselines}. Specifically, we evaluate on the test partition of \ctoc:\footnote{We used the test partition of \detokenize{data_2class} from the \ctoc dataset v1.0 (\texttt{https://huggingface.co/datasets/causalnlp/\newline corr2cause}), consisting of 2246 samples.} we do not use the training and validation partitions since we are not performing any fine-tuning. We then analyse the robustness of  \ourprompting to variable renaming and paraphrasing, including some natural story scenarios. Finally, we present a per partition performance analysis for two dataset partitions: per hypothesis and per number of variables. For all our results, we report two evaluation metrics: F1-score and accuracy, with the former being our main metric (similar to \citet{jin2024can}), since the dataset is imbalanced (15.23\% of the test set has positive label).

\begin{figure*}[t!]
    \centering
    \includegraphics[width=0.245\textwidth]{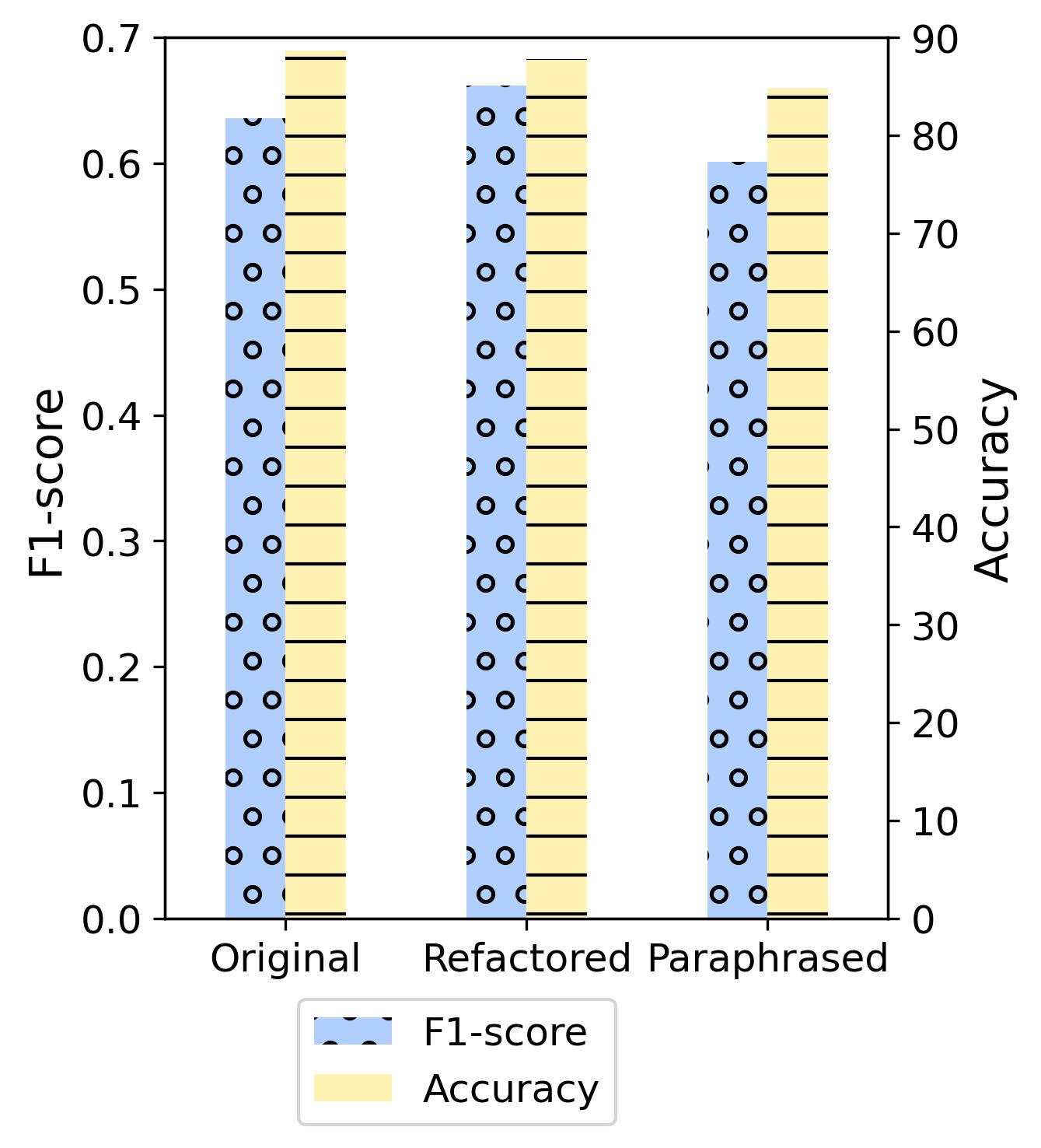}
    \hfill
    \includegraphics[width=0.75\textwidth]{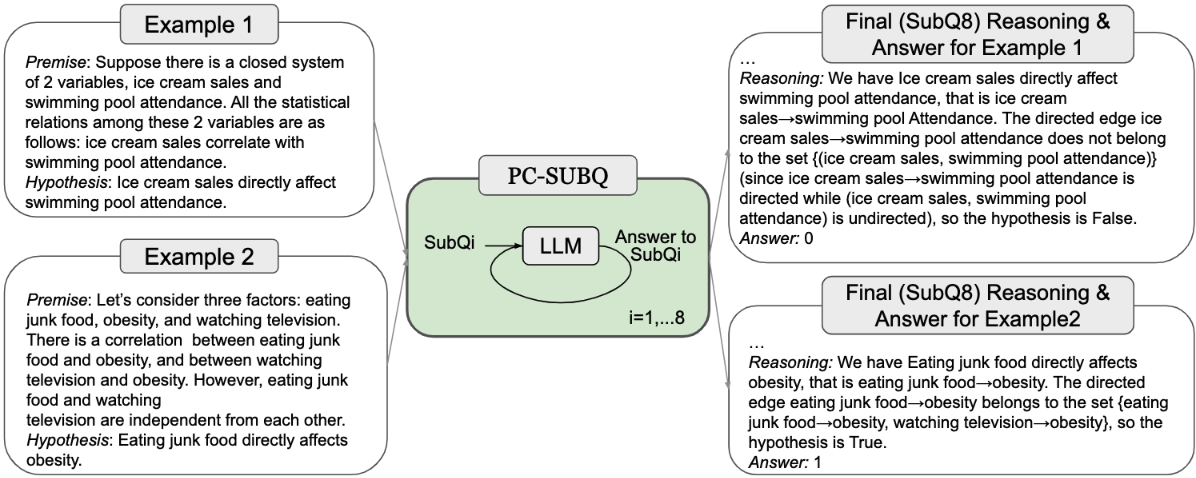}
    \caption{\textit{Left}: Performance using original, refactored and paraphrased \ourprompting prompts, showing robustness to these perturbations. \textit{Right}: \ourprompting reasoning and answers on two natural story examples. Output is correct even though natural story examples were never presented as few-shots.} 
    \label{fig:refactored_paraphrased}
\end{figure*} 

\paragraph{\ourprompting vs. other prompting strategies} We evaluate \ourprompting using five \llm{s}, namely Gemini Pro 1.0 \citep{geminiteam2023gemini}, Gemini Ultra 1.0 \citep{geminiteam2023gemini}, PaLM 2 of Large model size (PaLM 2 L) \citep{google2023palm2} and two GPT-family \llm{s} \citep{brown2020language, openai2023gpt}: GPT-3.5-turbo\footnote{\texttt{https://platform.openai.com/docs/models/\newline gpt-3-5-turbo}.} and GPT-4-turbo.\footnote{\texttt{https://platform.openai.com/docs/models/\newline gpt-4-turbo-and-gpt-4}.} Figure \ref{fig:results_all_models} illustrates the results, with each plot referring to one of the five \llm{s}.\footnote{GPT-4-turbo was only evaluated on a random 20\% sample of the dataset due to the induced high cost.}
The x-axis in all plots indicates the prompting strategy used: we compare \ourprompting with widely used prompting strategies, namely zero-shot, zero-shot \cotp \citep{kojima2023large}, zero-shot \cotp \pc, few-shot, few-shot \cotp \citep{wei2022chain}. Zero-shot \cotp \pc suggests to the \llm to apply the PC algorithm. Figure \ref{fig:baselines} includes specific details for each of these five strategies: for the two few-shot strategies we only show an indicative shot, while the six shots that we use in total in our experiments are provided in Appendix \ref{sec:appendix_baseline_shots}. "Always Majority" always outputs the majority (negative) class and as a result gives a very high accuracy but zero F1-score. We can observe that \ourprompting outperforms all the other prompting strategies, in terms of our main F1-score metric, across all five \llm{s}. In some cases the performance is improved by a large margin: remarkably, for PaLM 2, F1-score increases from $0.30$ (few-shot \cotp) to $0.64$ (\ourprompting) with a corresponding accuracy of $88.7\%$.

\vspace{-0.2cm}
\paragraph{Robustness analysis} In the following, we analyse the robustness of \ourprompting to causal query perturbations.
\citet{jin2024can} designed two robusteness tests, variable refactorization and paraphrasing, with the purpose of determining whether the high performance of their fine-tuned models was only due to overfitting on the \ctoc training data. For variable refactorization, they modified the variable names from $A, B, C, \ldots$ to $Z, Y, X, \dots$, while for paraphrasing, they changed the expression describing the causal relation to a semantically equivalent alternative (e.g., "causes" to "affects"). Running the fine-tuned models on refactored and paraphrased \ctoc test sets, they observed a drastic performance drop, showing that these models had not robustly learned the causal skill as they failed to generalize out-of-distribution. Overfitting is not an issue in our case, since we do not retrain the \llm{s}. Nevertheless, inspired by this robustness analysis, we check robustness by perturbing \ourprompting prompts such that they differ from the causal queries in the test dataset (see Appendix \ref{sec:appendix_perturbing} for an example). In particular, we modify \ourprompting few-shots and subquestions in terms of variable refactoring and paraphrasing, while keeping the test set (\texttt{[Premise]} and \texttt{[Hypothesis]}) fixed. 
Figure \ref{fig:refactored_paraphrased} (left) depicts the corresponding results using PaLM 2. Remarkably, the performance does not drop at all in case of refactoring and only slightly drops in case of paraphrasing.

Finally, we performed one more robustness test. One could imagine further scenarios of variable refactoring or paraphrasing by using natural instead of symbolic variables names, for instance, "Paracetamol is correlated with the capacity to empathize" instead of "$A$ correlates with $B$". 
For tasks involving natural variable names, notice that it may be non trivial to determine whether the \llm used formal causal reasoning or commonsense knowledge to reach its answer. Nevertheless, these natural examples could be considered more appealing and of wider interest.
Indeed, \citet{jin2024can} envision future extensions of the \ctoc dataset to various settings 
and provide a case study for how to generate more natural stories, with two such stories provided in their Appendix. Running PaLM 2 with \ourprompting (keeping the subquestions and few-shots fixed, containing only symbolic variable names) on these two natural stories produces correct reasoning traces and answers, as shown in Figure \ref{fig:refactored_paraphrased} (right). Only the output of the final \subq 8 is depicted, while the output of all subquestions can be found in Appendix \ref{sec:appendix_natural}. Note that in both examples the reasoning and final answers are correct even though we did not provide any natural story examples as few-shots.

\begin{figure}[!tbp]
    \begin{subfigure}{0.24\textwidth}
    \includegraphics[width=\textwidth]{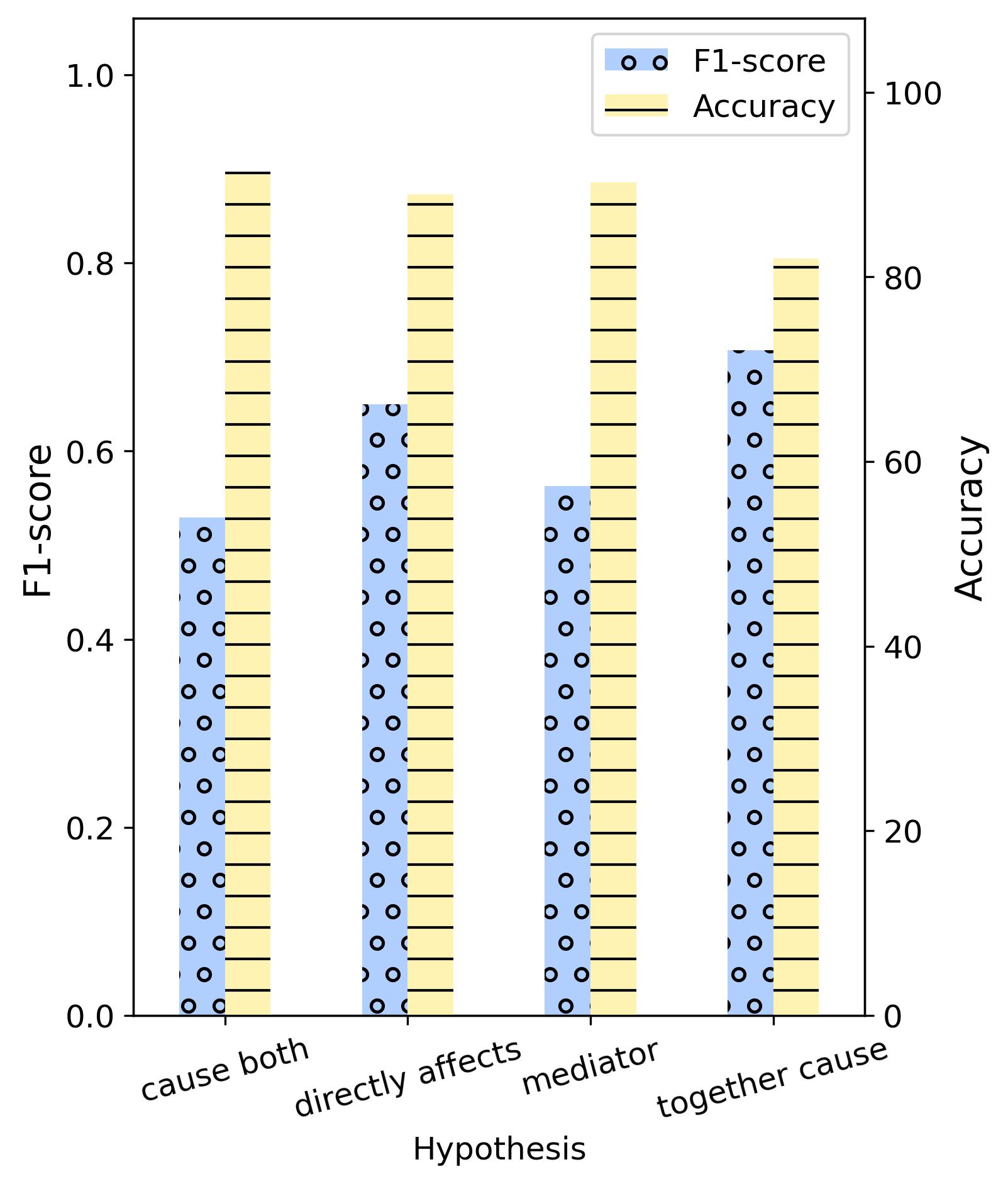}
    \caption{}
    \label{fig:results_hyp}
    \end{subfigure}%
    \begin{subfigure}{0.25\textwidth}
    \includegraphics[width=\textwidth]{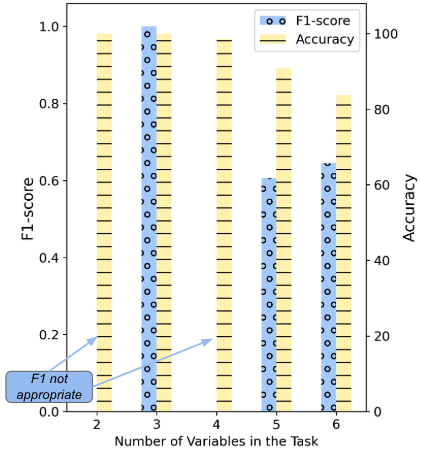}
    \caption{}
    \label{fig:results_numvar}
    \end{subfigure}
    \caption{\ourprompting performance (a) per hypothesis  and (b) per number of variables considered in the causal query, showing that more complex problems are harder to solve.}
    \label{fig:results_finegrained}
\end{figure}

\paragraph{Per hypothesis and per number of variables analysis} We additionally conducted an analysis partitioning the dataset into groups according to the hypothesis or according to the number of variables considered and determining the per group performance. Fig. \ref{fig:results_hyp} illustrates results per hypothesis for PaLM 2. Specifically, we partition into four groups depending on the expression used in the \textit{Hypothesis} to describe the causal relation. We do not observe large differences in the performance across the four groups, with the "directly affect" and "together cause" relations being better compared to the other two. Additionally, Fig. \ref{fig:results_numvar} depicts the fine-grained performance when partitioning according to the number of variables, $N=2,\ldots,6$, considered in the causal query. We can observe that the performance is much better (close to $100\%$ accuracy) when $N=2,3,4$, whereas there is a performance drop when the problem complexity increases ($N=5,6$). When $N=2$ or $N=4$, we do not report F1-score, as these subgroups only consist of records with negative validity label.   

\section{Discussion and Conclusion} \label{sec:discussion}

We introduced \ourprompting, a prompting strategy that enables \llm{s} to improve their performance on the task of inferring causal relationships from correlation statements.
\ourprompting decomposes the original \nlcd task into several \textit{fixed} subquestions, corresponding to the \pc algorithm steps, which compose the prompt chain given to an \llm.
Evaluated on \ctoc, \ourprompting outperforms a range of widely used prompting strategies across five \llm{s}. Furthermore, performance is robust to different variable instantiations or paraphrased causal queries. Remarkably, applied to examples with natural stories, \ourprompting produces correct reasoning traces and answers, even though few-shot demonstrations only contain examples with symbolic variable names.
Additionally, by guiding the \llm to reason through applying algorithmic steps, we gain more transparency and interpretability of the results: for instance, it is possible to trace which steps are responsible for an erroneous final answer by inspecting the output.
We provide such an example in Appendix \ref{sec:appendix_error_tracing} in which the erroneous final answer is only due to an error in the reasoning of the final \subq8 (confusing a directed edge with an undirected one), while the inferred graph of \subq7 and all previous reasoning traces and answers up to \subq7 are correct.
The proposed approach could be more generally applicable to address tasks that have a common underlying algorithm which can be applied to all examples to solve the task: each subquestion would then correspond to a step of this shared algorithm.
Finally, as with all prompting frameworks, there is no need to fine-tune a model using several thousands of annotated examples; instead, only a few (often less than 10) demonstrations are enough.

Our work has the following limitations. Current evaluations refer only to the output of the final \subq8, for which we have ground truth available.
Intermediate evaluations would be desirable in order to quantitatively assess intermediate answers, instead of only qualitatively inspecting the reasoning traces (as in Appendix \ref{sec:appendix_error_tracing}).
It would be particularly interesting to have evaluations for the output of \subq7, i.e., the inferred causal graph. Nevertheless, since the causal hypotheses of \ctoc cover causal relationships between \textit{every} pair of variables, current evaluations should be representative of the quality of the inferred graph.
Additionally, compared to few-shot prompting that requires only one \llm call, the proposed chain of prompts induces increased inference time due to the 16 \llm calls, two per subquestion for the reasoning and answer parts, needed. Future work should thus focus on making the strategy more efficient.
Furthermore, when the task complexity increases, as measured by the number of variables, performance is affected. This limitation might be mitigated for natural stories, since they often do not refer to more than 3 or 4 variables. 
Indeed, our results with such natural scenarios are very promising. Areas of future research could include constructing more \nlcd benchmarks with natural stories, beyond the two available ones, on which to employ \ourprompting.
The formal causal reasoning of \ourprompting could be further combined with the commonsense knowledge captured by \llm{s} for these scenarios.
Finally, an alternative path to tackle \nlcd could be via Tool Use, by interrupting the \llm reasoning process and calling a registered Python tool \citep{yao2023react, parisi2022talm}, with the \pc algorithm implemented. Notice that the \llm would still be required to understand the task in order to appropriately convert the input for the Python function and decide when to call it. We think of these two paths as complementary: as with other tasks, such as the ability to multiply two numbers or solve a math or physics problem, it is desirable for the \llm to have certain skills and not solely rely on external tools, which also may not be always available.

\typeout{}
\bibliography{bib}

\begin{thebibliography}{41}
\providecommand{\natexlab}[1]{#1}
\providecommand{\url}[1]{\texttt{#1}}
\expandafter\ifx\csname urlstyle\endcsname\relax
  \providecommand{\doi}[1]{doi: #1}\else
  \providecommand{\doi}{doi: \begingroup \urlstyle{rm}\Url}\fi

\bibitem[Abdulaal et~al.(2024)Abdulaal, adamos hadjivasiliou, Montana-Brown,
  He, Ijishakin, Drobnjak, Castro, and Alexander]{abdulaal2024causal}
Abdulaal, A., adamos hadjivasiliou, Montana-Brown, N., He, T., Ijishakin, A.,
  Drobnjak, I., Castro, D.~C., and Alexander, D.~C.
\newblock Causal modelling agents: Causal graph discovery through synergising
  metadata- and data-driven reasoning.
\newblock In \emph{The Twelfth International Conference on Learning
  Representations}, 2024.

\bibitem[Authors(2023)]{big-bench2023beyond}
Authors, B.
\newblock Beyond the imitation game: Quantifying and extrapolating the
  capabilities of language models.
\newblock \emph{Transactions on Machine Learning Research}, 2023.
\newblock ISSN 2835-8856.

\bibitem[Brown et~al.(2020)Brown, Mann, Ryder, Subbiah, Kaplan, Dhariwal,
  Neelakantan, Shyam, Sastry, Askell, et~al.]{brown2020language}
Brown, T., Mann, B., Ryder, N., Subbiah, M., Kaplan, J.~D., Dhariwal, P.,
  Neelakantan, A., Shyam, P., Sastry, G., Askell, A., et~al.
\newblock Language models are few-shot learners.
\newblock In \emph{Advances in Neural Information Processing Systems}, pp.\
  1877--1901, 2020.

\bibitem[Chickering(2002)]{chickering2002learning}
Chickering, D.~M.
\newblock Learning equivalence classes of {B}ayesian-network structures.
\newblock \emph{Journal of Machine Learning Research}, 2:\penalty0 445--498,
  2002.

\bibitem[Cohrs et~al.(2024)Cohrs, Diaz, Sitokonstantinou, Varando, and
  Camps-Valls]{cohrs2023large}
Cohrs, K.-H., Diaz, E., Sitokonstantinou, V., Varando, G., and Camps-Valls, G.
\newblock Large language models for constrained-based causal discovery.
\newblock In \emph{{AAAI} 2024 Workshop on ``Are Large Language Models Simply
  Causal Parrots?''}, 2024.

\bibitem[Dua et~al.(2022)Dua, Gupta, Singh, and Gardner]{dua2022successive}
Dua, D., Gupta, S., Singh, S., and Gardner, M.
\newblock Successive prompting for decomposing complex questions.
\newblock In \emph{Conference on Empirical Methods in Natural Language
  Processing}, pp.\  1251--1265, 2022.

\bibitem[{Gemini Team Google}(2023)]{geminiteam2023gemini}
{Gemini Team Google}.
\newblock Gemini: A family of highly capable multimodal models.
\newblock \emph{arXiv preprint arXiv:2312.11805}, 2023.

\bibitem[Google~and et~al.(2023)Google~and, Dai, Firat, Johnson, Lepikhin,
  Passos, Shakeri, Taropa, Bailey, Chen, Chu, Clark, Shafey, Huang,
  Meier-Hellstern, Mishra, Moreira, Omernick, Robinson, Ruder, Tay, Xiao, Xu,
  Zhang, Abrego, Ahn, Austin, Barham, Botha, Bradbury, Brahma, Brooks, Catasta,
  Cheng, Cherry, Choquette-Choo, Chowdhery, Crepy, Dave, Dehghani, Dev, Devlin,
  Díaz, Du, Dyer, Feinberg, Feng, Fienber, Freitag, Garcia, Gehrmann,
  Gonzalez, Gur-Ari, Hand, Hashemi, Hou, Howland, Hu, Hui, Hurwitz, Isard,
  Ittycheriah, Jagielski, Jia, Kenealy, Krikun, Kudugunta, Lan, Lee, Lee, Li,
  Li, Li, Li, Li, Lim, Lin, Liu, Liu, Maggioni, Mahendru, Maynez, Misra,
  Moussalem, Nado, Nham, Ni, Nystrom, Parrish, Pellat, Polacek, Polozov, Pope,
  Qiao, Reif, Richter, Riley, Ros, Roy, Saeta, Samuel, Shelby, Slone, Smilkov,
  So, Sohn, Tokumine, Valter, Vasudevan, Vodrahalli, Wang, Wang, Wang, Wang,
  Wieting, Wu, Xu, Xu, Xue, Yin, Yu, Zhang, Zheng, Zheng, Zhou, Zhou, Petrov,
  and Wu]{google2023palm2}
Google~and, R.~A., Dai, A.~M., Firat, O., Johnson, M., Lepikhin, D., Passos,
  A., Shakeri, S., Taropa, E., Bailey, P., Chen, Z., Chu, E., Clark, J.~H.,
  Shafey, L.~E., Huang, Y., Meier-Hellstern, K., Mishra, G., Moreira, E.,
  Omernick, M., Robinson, K., Ruder, S., Tay, Y., Xiao, K., Xu, Y., Zhang, Y.,
  Abrego, G.~H., Ahn, J., Austin, J., Barham, P., Botha, J., Bradbury, J.,
  Brahma, S., Brooks, K., Catasta, M., Cheng, Y., Cherry, C., Choquette-Choo,
  C.~A., Chowdhery, A., Crepy, C., Dave, S., Dehghani, M., Dev, S., Devlin, J.,
  Díaz, M., Du, N., Dyer, E., Feinberg, V., Feng, F., Fienber, V., Freitag,
  M., Garcia, X., Gehrmann, S., Gonzalez, L., Gur-Ari, G., Hand, S., Hashemi,
  H., Hou, L., Howland, J., Hu, A., Hui, J., Hurwitz, J., Isard, M.,
  Ittycheriah, A., Jagielski, M., Jia, W., Kenealy, K., Krikun, M., Kudugunta,
  S., Lan, C., Lee, K., Lee, B., Li, E., Li, M., Li, W., Li, Y., Li, J., Lim,
  H., Lin, H., Liu, Z., Liu, F., Maggioni, M., Mahendru, A., Maynez, J., Misra,
  V., Moussalem, M., Nado, Z., Nham, J., Ni, E., Nystrom, A., Parrish, A.,
  Pellat, M., Polacek, M., Polozov, A., Pope, R., Qiao, S., Reif, E., Richter,
  B., Riley, P., Ros, A.~C., Roy, A., Saeta, B., Samuel, R., Shelby, R., Slone,
  A., Smilkov, D., So, D.~R., Sohn, D., Tokumine, S., Valter, D., Vasudevan,
  V., Vodrahalli, K., Wang, X., Wang, P., Wang, Z., Wang, T., Wieting, J., Wu,
  Y., Xu, K., Xu, Y., Xue, L., Yin, P., Yu, J., Zhang, Q., Zheng, S., Zheng,
  C., Zhou, W., Zhou, D., Petrov, S., and Wu, Y.
\newblock Pa{LM} 2 technical report.
\newblock \emph{arXiv preprint arXiv:2305.10403}, 2023.

\bibitem[Hoyer et~al.(2008)Hoyer, Janzing, Mooij, Peters, and
  Sch{\"o}lkopf]{hoyer2008nonlinear}
Hoyer, P., Janzing, D., Mooij, J.~M., Peters, J., and Sch{\"o}lkopf, B.
\newblock Nonlinear causal discovery with additive noise models.
\newblock \emph{Advances in Neural Information Processing Systems}, 2008.

\bibitem[Huang \& Chang(2023)Huang and Chang]{huang2022towards}
Huang, J. and Chang, K. C.-C.
\newblock Towards reasoning in large language models: A survey.
\newblock In \emph{Findings of the Association for Computational Linguistics:
  {ACL} 2023}, pp.\  1049--1065, 2023.

\bibitem[Jin et~al.(2023)Jin, Chen, Leeb, Gresele, Kamal, Zhiheng, Blin,
  Adauto, Kleiman-Weiner, Sachan, and Sch{\"o}lkopf]{jin2023cladder}
Jin, Z., Chen, Y., Leeb, F., Gresele, L., Kamal, O., Zhiheng, L., Blin, K.,
  Adauto, F.~G., Kleiman-Weiner, M., Sachan, M., and Sch{\"o}lkopf, B.
\newblock {CL}adder: Assessing causal reasoning in language models.
\newblock In \emph{Advances in Neural Information Processing Systems}, 2023.

\bibitem[Jin et~al.(2024)Jin, Liu, Lyu, Poff, Sachan, Mihalcea, Diab, and
  Sch{\"o}lkopf]{jin2024can}
Jin, Z., Liu, J., Lyu, Z., Poff, S., Sachan, M., Mihalcea, R., Diab, M.~T., and
  Sch{\"o}lkopf, B.
\newblock Can large language models infer causation from correlation?
\newblock In \emph{International Conference on Learning Representations}, 2024.

\bibitem[Jiralerspong et~al.(2024)Jiralerspong, Chen, More, Shah, and
  Bengio]{jiralerspong2024efficient}
Jiralerspong, T., Chen, X., More, Y., Shah, V., and Bengio, Y.
\newblock Efficient causal graph discovery using large language models.
\newblock \emph{arXiv preprint arXiv:2402.01207}, 2024.

\bibitem[Khot et~al.(2023)Khot, Trivedi, Finlayson, Fu, Richardson, Clark, and
  Sabharwal]{khot2023decomposed}
Khot, T., Trivedi, H., Finlayson, M., Fu, Y., Richardson, K., Clark, P., and
  Sabharwal, A.
\newblock Decomposed prompting: A modular approach for solving complex tasks.
\newblock In \emph{International Conference on Learning Representations}, 2023.

\bibitem[K{\i}c{\i}man et~al.(2023)K{\i}c{\i}man, Ness, Sharma, and
  Tan]{kiciman2023causal}
K{\i}c{\i}man, E., Ness, R., Sharma, A., and Tan, C.
\newblock Causal reasoning and large language models: Opening a new frontier
  for causality.
\newblock \emph{arXiv preprint arXiv:2305.00050}, 2023.

\bibitem[Kojima et~al.(2022)Kojima, Gu, Reid, Matsuo, and
  Iwasawa]{kojima2023large}
Kojima, T., Gu, S.~S., Reid, M., Matsuo, Y., and Iwasawa, Y.
\newblock Large language models are zero-shot reasoners.
\newblock 2022.

\bibitem[Long et~al.(2022)Long, Schuster, and Pich{\'e}]{long2022can}
Long, S., Schuster, T., and Pich{\'e}, A.
\newblock Can large language models build causal graphs?
\newblock In \emph{{NeurIPS} 2022 Workshop on Causal Machine Learning for
  Real-World Impact}, 2022.

\bibitem[Long et~al.(2023)Long, Pich{\'e}, Zantedeschi, Schuster, and
  Drouin]{long2023causal}
Long, S., Pich{\'e}, A., Zantedeschi, V., Schuster, T., and Drouin, A.
\newblock Causal discovery with language models as imperfect experts.
\newblock In \emph{{ICML} Workshop on Structured Probabilistic Inference \&
  Generative Modeling}, 2023.

\bibitem[Markeeva et~al.(2024)Markeeva, McLeish, Ibarz, Bounsi, Kozlova,
  Vitvitskyi, Blundell, Goldstein, Schwarzschild, and
  Veli{\v{c}}kovi{\'c}]{markeeva2024clrs}
Markeeva, L., McLeish, S., Ibarz, B., Bounsi, W., Kozlova, O., Vitvitskyi, A.,
  Blundell, C., Goldstein, T., Schwarzschild, A., and Veli{\v{c}}kovi{\'c}, P.
\newblock The clrs-text algorithmic reasoning language benchmark.
\newblock \emph{arXiv preprint arXiv:2406.04229}, 2024.

\bibitem[Nye et~al.(2021)Nye, Andreassen, Gur-Ari, Michalewski, Austin, Bieber,
  Dohan, Lewkowycz, Bosma, Luan, et~al.]{nye2021show}
Nye, M., Andreassen, A.~J., Gur-Ari, G., Michalewski, H., Austin, J., Bieber,
  D., Dohan, D., Lewkowycz, A., Bosma, M., Luan, D., et~al.
\newblock Show your work: Scratchpads for intermediate computation with
  language models.
\newblock \emph{arXiv preprint arXiv:2112.00114}, 2021.

\bibitem[OpenAI(2023)]{openai2023gpt}
OpenAI.
\newblock {GPT}-4 technical report.
\newblock \emph{arXiv preprint arXiv:2303.08774}, 2023.

\bibitem[Parisi et~al.(2022)Parisi, Zhao, and Fiedel]{parisi2022talm}
Parisi, A., Zhao, Y., and Fiedel, N.
\newblock Talm: Tool augmented language models, 2022.

\bibitem[Pearl(2000)]{pearl2000causality}
Pearl, J.
\newblock \emph{Causality: Models, Reasoning, and Inference}.
\newblock Cambridge University Press, 2000.

\bibitem[Pearl \& Mackenzie(2018)Pearl and Mackenzie]{pearl2018book}
Pearl, J. and Mackenzie, D.
\newblock \emph{The Book of Why: The New Science of Cause and Effect}.
\newblock Basic Books, 2018.

\bibitem[Peters et~al.(2014)Peters, Mooij, Janzing, and
  Sch{\"o}lkopf]{peters2014causal}
Peters, J., Mooij, J.~M., Janzing, D., and Sch{\"o}lkopf, B.
\newblock Causal discovery with continuous additive noise models.
\newblock \emph{Journal of Machine Learning Research}, 15:\penalty0 2009--2053,
  2014.

\bibitem[Qiao et~al.(2023)Qiao, Ou, Zhang, Chen, Yao, Deng, Tan, Huang, and
  Chen]{qiao2023reasoning}
Qiao, S., Ou, Y., Zhang, N., Chen, X., Yao, Y., Deng, S., Tan, C., Huang, F.,
  and Chen, H.
\newblock Reasoning with language model prompting: A survey.
\newblock In \emph{Annual Meeting of the Association for Computational
  Linguistics (Volume 1: Long Papers)}, pp.\  5368--5393, 2023.

\bibitem[Rae et~al.(2021)Rae, Borgeaud, Cai, Millican, Hoffmann, Song,
  Aslanides, Henderson, Ring, Young, et~al.]{rae2021scaling}
Rae, J.~W., Borgeaud, S., Cai, T., Millican, K., Hoffmann, J., Song, F.,
  Aslanides, J., Henderson, S., Ring, R., Young, S., et~al.
\newblock Scaling language models: Methods, analysis \& insights from training
  gopher.
\newblock \emph{arXiv preprint arXiv:2112.11446}, 2021.

\bibitem[Sel et~al.(2023)Sel, Al-Tawaha, Khattar, Wang, Jia, and
  Jin]{sel2023algorithm}
Sel, B., Al-Tawaha, A., Khattar, V., Wang, L., Jia, R., and Jin, M.
\newblock Algorithm of thoughts: Enhancing exploration of ideas in large
  language models.
\newblock \emph{CoRR}, abs/2308.10379, 2023.
\newblock URL \url{https://doi.org/10.48550/arXiv.2308.10379}.

\bibitem[Shimizu et~al.(2006)Shimizu, Hoyer, Hyv{\"a}rinen, Kerminen, and
  Jordan]{shimizu2006linear}
Shimizu, S., Hoyer, P.~O., Hyv{\"a}rinen, A., Kerminen, A., and Jordan, M.
\newblock A linear non-{G}aussian acyclic model for causal discovery.
\newblock \emph{Journal of Machine Learning Research}, 7:\penalty0 2003--2030,
  2006.

\bibitem[Spirtes et~al.(2000)Spirtes, Glymour, and
  Scheines]{spirtes2000causation}
Spirtes, P., Glymour, C.~N., and Scheines, R.
\newblock \emph{Causation, Prediction, and Search}.
\newblock MIT Press, 2000.

\bibitem[Sun et~al.(2023)Sun, Zheng, Xie, Liu, Chu, Qiu, Xu, Ding, Li, Geng,
  et~al.]{sun2023survey}
Sun, J., Zheng, C., Xie, E., Liu, Z., Chu, R., Qiu, J., Xu, J., Ding, M., Li,
  H., Geng, M., et~al.
\newblock A survey of reasoning with foundation models.
\newblock \emph{arXiv preprint arXiv:2312.11562}, 2023.

\bibitem[Tu et~al.(2023)Tu, Ma, and Zhang]{tu2023causal}
Tu, R., Ma, C., and Zhang, C.
\newblock Causal-discovery performance of {C}hat{GPT} in the context of
  neuropathic pain diagnosis.
\newblock \emph{arXiv preprint arXiv:2301.13819}, 2023.

\bibitem[Vashishtha et~al.(2023)Vashishtha, Reddy, Kumar, Bachu,
  Balasubramanian, and Sharma]{vashishtha2023causal}
Vashishtha, A., Reddy, A.~G., Kumar, A., Bachu, S., Balasubramanian, V.~N., and
  Sharma, A.
\newblock Causal inference using llm-guided discovery.
\newblock \emph{arXiv preprint arXiv:2310.15117}, 2023.

\bibitem[Veli{\v{c}}kovi{\'c} et~al.(2022)Veli{\v{c}}kovi{\'c}, Badia, Budden,
  Pascanu, Banino, Dashevskiy, Hadsell, and Blundell]{velivckovic2022clrs}
Veli{\v{c}}kovi{\'c}, P., Badia, A.~P., Budden, D., Pascanu, R., Banino, A.,
  Dashevskiy, M., Hadsell, R., and Blundell, C.
\newblock The clrs algorithmic reasoning benchmark.
\newblock In \emph{International Conference on Machine Learning}, pp.\
  22084--22102. PMLR, 2022.

\bibitem[Wei et~al.(2022)Wei, Wang, Schuurmans, Bosma, Xia, Chi, Le, Zhou,
  et~al.]{wei2022chain}
Wei, J., Wang, X., Schuurmans, D., Bosma, M., Xia, F., Chi, E., Le, Q.~V.,
  Zhou, D., et~al.
\newblock Chain-of-thought prompting elicits reasoning in large language
  models.
\newblock pp.\  24824--24837, 2022.

\bibitem[Willig et~al.(2022)Willig, Ze{\v{c}}evi{\'c}, Dhami, and
  Kersting]{willig2022can}
Willig, M., Ze{\v{c}}evi{\'c}, M., Dhami, D.~S., and Kersting, K.
\newblock Can foundation models talk causality?
\newblock In \emph{{UAI} Workshop on Causal Representation Learning}, 2022.

\bibitem[Yao et~al.(2023)Yao, Zhao, Yu, Du, Shafran, Narasimhan, and
  Cao]{yao2023react}
Yao, S., Zhao, J., Yu, D., Du, N., Shafran, I., Narasimhan, K., and Cao, Y.
\newblock React: Synergizing reasoning and acting in language models.
\newblock In \emph{International Conference on Learning Representations}, 2023.

\bibitem[Ze{\v{c}}evi{\'c} et~al.(2023)Ze{\v{c}}evi{\'c}, Willig, Dhami, and
  Kersting]{zevcevic2023causal}
Ze{\v{c}}evi{\'c}, M., Willig, M., Dhami, D.~S., and Kersting, K.
\newblock Causal parrots: Large language models may talk causality but are not
  causal.
\newblock \emph{Transactions on Machine Learning Research}, 2023.

\bibitem[Zhang et~al.(2023)Zhang, Bauer, Bennett, Gao, Gong, Hilmkil, Jennings,
  Ma, Minka, Pawlowski, and Vaughan]{zhang2023understanding}
Zhang, C., Bauer, S., Bennett, P., Gao, J., Gong, W., Hilmkil, A., Jennings,
  J., Ma, C., Minka, T., Pawlowski, N., and Vaughan, J.
\newblock Understanding causality with large language models: Feasibility and
  opportunities.
\newblock \emph{arXiv preprint arXiv:2304.05524}, 2023.

\bibitem[Zhou et~al.(2023{\natexlab{a}})Zhou, Sch{\"a}rli, Hou, Wei, Scales,
  Wang, Schuurmans, Cui, Bousquet, Le, and Chi]{zhou2023least}
Zhou, D., Sch{\"a}rli, N., Hou, L., Wei, J., Scales, N., Wang, X., Schuurmans,
  D., Cui, C., Bousquet, O., Le, Q., and Chi, E.~H.
\newblock Least-to-most prompting enables complex reasoning in large language
  models.
\newblock In \emph{International Conference on Learning Representations},
  2023{\natexlab{a}}.

\bibitem[Zhou et~al.(2023{\natexlab{b}})Zhou, Nova, Courville, Larochelle,
  Neyshabur, and Sedghi]{zhou2023teaching}
Zhou, H., Nova, A., Courville, A., Larochelle, H., Neyshabur, B., and Sedghi,
  H.
\newblock Teaching algorithmic reasoning via in-context learning.
\newblock In \emph{International Conference on Learning Representations},
  2023{\natexlab{b}}.

\end{thebibliography}
\bibliographystyle{icml2025}

\newpage
\appendix
\onecolumn

\section{\ourprompting: Few-shot Examples per Subquestion}\label{sec:appendix_subquestions_shots}

Tables \ref{tab:appendix-subq1}-\ref{tab:appendix-subq8-c} provide all few-shot examples prepended to every subquestion of \ourprompting. These can be used together with the five off-the-shelf \llm{s} to reproduce the performance of \ourprompting reported in Fig. \ref{fig:results_all_models}. 
In the following, we provide some additional clarifications for Tables \ref{tab:appendix-subq1}-\ref{tab:appendix-subq8-c}:
\begin{itemize}
    \item $(\cdot, \cdot)$ denotes an undirected edge, while $\cdot \rightarrow \cdot$ represents a directed one.
    \item For symbols ``$\Rightarrow$'' and ``$\rightarrow$'', we use ``=>'' and ``->'', respectively, in the prompts, and for variable names just capital letters A, B, C, \ldots
    \item We perform two calls per subquestion: the first call extracts the \textit{Reasoning} and is using exactly the prompt included in each subquestion's Table below.\footnote{The Reasoning consists of the output of the \llm until the ``Answer:'' string is met.} The second call extracts the \textit{Answer} and is using the previous prompt augmented with the \textit{Reasoning} output of the first call and the ``Answer:'' string appended in the end.
    \item No other information is passed from previous subquestions to the prompt of the next ones apart from what is included in every subquestion's Table below. For instance (see Table \ref{tab:appendix-subq3}), \subq3 only needs the \texttt{[Answer to \subq2]}. All other previous history, such as the question, reasoning and answer of \subq1 or the question and reasoning of \subq2, is not propagated.
\end{itemize}

\begin{table}[h!]
    \caption{    
    Few-shot exemplars for \subq1 followed by \subq1 at the end.
    }
    \centering
    \small
    \begin{tabular}{p{\linewidth}}
        \toprule
        \textbf{Question}: Premise: Suppose there is a closed system of 3 variables, $A$, $B$ and $C$. All the statistical relations among these 3 variables are as follows: $A$ correlates with $C$. $B$ correlates with $C$. However, $A$ is independent of $B$. Can you initialize the \pc algorithm with a fully connected undirected graph? \\
        \vspace{-1.4mm}
        \textbf{Reasoning}: Since our variables are $A$,$B$,$C$ $\Rightarrow$ the initial fully connected undirected graph is $(A,B), (A,C), (B,C)$. \\
        \vspace{-1.4mm}
         \textbf{Answer}: $(A,B), (A,C), (B,C)$ \\
         \vspace{1mm}
         \textbf{Question}: \texttt{[Premise]}. Can you initialize the PC algorithm with a fully connected undirected graph? \\
         \vspace{-1.4mm}
        \textbf{Reasoning}: \\ 
        \bottomrule
    \end{tabular}
    \label{tab:appendix-subq1}
\end{table}

\begin{table}[h!]
    \caption{    
    Few-shot exemplars for \subq2 followed by \subq2 at the end.
    }
    \centering
    \small
    \begin{tabular}{p{\linewidth}}
        \toprule
        \textbf{Question}: Premise: Suppose there is a closed system of 3 variables, $A$, $B$ and $C$. All the statistical relations among these 3 variables are as follows: $A$ correlates with $C$. $B$ correlates with $C$. However, $A$ is independent of $B$. Given the initial fully connected graph: $(A,B), (A,C), (B,C)$. Can you apply the first step of the \pc algorithm to infer the undirected graph between the variables? \\
         \vspace{-1.4mm}
         \textbf{Reasoning}: \\
         1. We start with the given fully connected graph: $(A,B), (A,C), (B,C)$.\newline 2. We then check all conditional independencies and remove edges appropriately. In our case: $A$ is independent of $B$ $\Rightarrow$ $(A,B)$ is removed $\Rightarrow$ the graph after this step is $(A,C), (B,C)$. \\
         \vspace{-1.4mm}
         3. The final undirected graph is: $(A,C), (B,C)$. \\
         \textbf{Answer}: $(A,C), (B,C)$ \\
         \vspace{1mm}
         \textbf{Question}: \texttt{[Premise]}. Given the initial fully connected graph: \texttt{[Answer to SubQ1]} \\
         Can you apply the first step of the PC algorithm to infer the undirected graph between the variables? \\
         \vspace{-1.4mm}
        \textbf{Reasoning}: \\ 
        \bottomrule
    \end{tabular}
    \label{tab:appendix-subq2}
\end{table}


\begin{table}[h!]
    \caption{    
    Few-shot exemplars for \subq3 followed by \subq3 at the end.
    }
    \centering
    \small
    \begin{tabular}{p{\linewidth}}
        \toprule
        \textbf{Question}: Given the undirected graph: $(A,B), (A,D), (A,E), (B,C)$. Can you find all paths of length 2? \\
        \vspace{-1.4mm}
        \textbf{Reasoning}: We go through all unordered pairs of edges in the undirected graph $(A,B), (A,D), (A,E), (B,C)$ above and find paths of length 2: \\
        Since the edges $(A,B)$ and $(A,D)$ share a common node $A$ $\Rightarrow$ $(B,A,D)$ is a path of length 2. \\
        Since the edges $(A,B)$ and $(A,E)$ share a common node $A$ $\Rightarrow$ $(B,A,E)$ is a path of length 2. \\
        Since the edges $(A,B)$ and $(B,C)$ share a common node $B$ $\Rightarrow$ $(A,B,C)$ is a path of length 2. \\
        Since the edges $(A,D)$ and $(A,E)$ share a common node $A$ $\Rightarrow$ $(D,A,E)$ is a path of length 2. \\
        Since the edges $(A,D)$ and $(B,C)$ do not have a common node, they do not form a path of length 2. \\
        Since the edges $(A,E)$ and $(B,C)$ do not have a common node, they do not form a path of length 2. \\
        \vspace{-1.4mm}
        \textbf{Answer}: $(B,A,D), (B,A,E), (A,B,C), (D,A,E)$ \\
        \vspace{1mm}
        \textbf{Question}: Given the undirected graph: $(C,D)$. Can you find all paths of length 2? \\
        \vspace{-1.4mm}
        \textbf{Reasoning}: We go through all unordered pairs of edges in the undirected graph $(C,D)$ above and find paths of length 2: \\
        Since there is only one edge, there are no paths of length 2. \\
        \vspace{-1.4mm}
        \textbf{Answer}: No paths of length 2 found. \\
        \vspace{1mm} 
        \textbf{Question}: Given the undirected graph: $(A,C), (A,D), (B,C)$. Can you find all paths of length 2? \\
        \vspace{-1.4mm}
        \textbf{Reasoning}: We go through all unordered pairs of edges in the undirected graph $(A,C), (A,D), (B,C)$ above and find paths of length 2: \newline
        Since the edges $(A,C)$ and $(A,D)$ share a common node $A$ $\Rightarrow$ $(C,A,D)$ is a path of length 2. \\
        Since the edges $(A,C)$ and $(B,C)$ share a common node $C$ $\Rightarrow$ $(A,C,B)$ is a path of length 2. \\
        Since the edges $(A,D)$ and $(B,C)$ do not have a common node, they do not form a path of length 2. \\
        \vspace{-1.4mm}
        \textbf{Answer}: $(C,A,D)$, $(A,C,B)$ \\
        \vspace{1mm} 
        \textbf{Question}:  Given the undirected graph: \texttt{[Answer to SubQ2]} Can you find all paths of length 2? \\
        \vspace{-1.4mm}
        \textbf{Reasoning}: \\ 
        \bottomrule
    \end{tabular}
    \label{tab:appendix-subq3}
\end{table}

\begin{table}[h!]
    \caption{    
    Few-shot exemplars for \subq4 followed by \subq4 at the end.
    }
    \centering
    \small
    \begin{tabular}{p{\linewidth}}
        \toprule
        \textbf{Question}: Given the paths of length 2: $(B,A,C), (A,B,C), (A,C,B)$ of the undirected graph: $(A,B), (A,C), (B,C)$. Can you find all possible v-structures? \\
        \vspace{-1.4mm}
        \textbf{Reasoning}: We go through all paths of length 2 and check if there is an edge connecting the start and end of the path in the undirected graph: $(A,B)$, $(A,C)$, $(B,C)$. \\
        For path $(B,A,C)$: $(B,C)$ belongs to the set of edges $\{(A,B), (A,C), (B,C)\}$ $\Rightarrow$ $(B,A,C)$ is not a v-structure. \\
        For path $(A,B,C)$: $(A,C)$ belongs to the set of edges $\{(A,B), (A,C), (B,C)\}$ $\Rightarrow$ $(A,B,C)$ is not a v-structure. \\
        For path $(A,C,B)$: $(A,B)$ belongs to the set of edges $\{(A,B), (A,C), (B,C)\}$ $\Rightarrow$ $(A,C,B)$ is not a v-structure. \\
        \vspace{-1.4mm}
        \textbf{Answer}: No possible v-structures found \\
        \vspace{1mm}
        \textbf{Question}: Given the paths of length 2: $(A,C,B)$ of the undirected graph: $(A,C)$, $(B,C)$. Can you find all possible v-structures? \\
        \vspace{-1.4mm}
        \textbf{Reasoning}: We go through all paths of length 2 and check if there is an edge connecting the start and end of the path in the undirected graph: $(A,C)$, $(B,C)$. \\
        For path $(A,C,B)$: $(A,B)$ does not belong to the set of edges $\{(A,C), (B,C)\}$ $\Rightarrow$ $(A,C,B)$ is a possible v-structure. \\
        \vspace{-1.4mm}
        \textbf{Answer}: $(A,C,B)$ \\
        \vspace{1mm}
        \textbf{Question}: Given the paths of length 2: No paths of length 2 found of the undirected graph: $(C,D)$. Can you find all possible v-structures? \\
        \vspace{-1.4mm}
        \textbf{Reasoning}: No paths of length 2 found $\Rightarrow$ No possible v-structures found. \\
        \vspace{-1.4mm}
        \textbf{Answer}: No possible v-structures found \\
        \vspace{1mm}
        \textbf{Question}: Given the paths of length 2: $(B,A,D), (A,B,D), (A,D,B), (A,D,C), (B,D,C)$ of the undirected graph: $(A,B), (A,D), (B,D), (C,D)$. Can you find all possible v-structures? \\
        \vspace{-1.4mm}
        \textbf{Reasoning}: We go through all paths of length 2 and check if there is an edge connecting the start and end of the path in the undirected graph: $(A,B), (A,D), (B,D), (C,D)$. \\
        For path $(B,A,D)$: $(B,D)$ belongs to the set of edges $\{(A,B), (A,D), (B,D), (C,D)\}$ $\Rightarrow$ $(B,A,D)$ is not a v-structure. \\
        For path $(A,B,D)$: $(A,D)$ belongs to the set of edges $\{(A,B), (A,D), (B,D), (C,D)\}$ $\Rightarrow$ $(A,B,D)$ is not a v-structure. \\
        For path $(A,D,B)$: $(A,B)$ belongs to the set of edges $\{(A,B), (A,D), (B,D), (C,D)\}$ $\Rightarrow$ $(A,D,B)$ is not a v-structure. \\
        For path $(A,D,C)$: $(A,C)$ does not belong to the set of edges $\{(A,B), (A,D), (B,D), (C,D)\}$ $\Rightarrow$ $(A,D,C)$ is a possible v-structure. \\
        For path $(B,D,C)$: $(B,C)$ does not belong to the set of edges $\{(A,B), (A,D), (B,D), (C,D)\}$ $\Rightarrow$ $(B,D,C)$ is a possible v-structure. \\
        \vspace{-1.4mm}
        \textbf{Answer}: $(A,D,C), (B,D,C)$ \\
        \vspace{1mm}
        \textbf{Question}: Given the paths of length 2: \texttt{[Answer to SubQ3]} of the undirected graph: \texttt{[Answer to SubQ2]} Can you find all possible v-structures of the undirected graph? \\
        \vspace{-1.4mm}
        \textbf{Reasoning}: \\ 
        \bottomrule
    \end{tabular}
    \label{tab:appendix-subq4}
\end{table}

\begin{table}[h!]
    \caption{    
    Few-shot exemplars for \subq5 followed by \subq5 at the end.
    }
    \centering
    \small
    \begin{tabular}{p{\linewidth}}
        \toprule
        \textbf{Question}: Premise: Suppose there is a closed system of 4 variables, $A$, $B$, $C$ and $D$. All the statistical relations among these 3 variables are as follows: $A$ correlates with $B$. $A$ correlates with $D$. $B$ correlates with $D$. $C$ correlates with $D$. However, $A$ is independent of $C$. $A$ and $C$ are independent given $B$. $B$ is independent of $C$. $B$ and $C$ are independent given $A$.
        Given the possible v-structures from the previous step: 
        $(A,D,C)$, $(B,D,C)$. Can you apply the second step of the \pc algorithm to orient these v-structures? \\
        \vspace{-1.4mm}
        \textbf{Reasoning}: We first go through all possible v-structures: \\
        \vspace{-1.4mm}
        $(A,D,C)$ is a possible v-structure: all conditional independences between $A$ and $C$ are: $A$ is independent of $C$. $A$ and $C$ are independent given $B$. Since the middle variable $D$ is not in any conditioning set that makes $A$ and $C$ independent $\Rightarrow$ $(A,D,C)$ form a v-structure $\Rightarrow$ we orient the arrows towards the middle node $D$: $A \rightarrow D$, $C \rightarrow D$. \\
        $(B,D,C)$ is a possible v-structure: all conditional independences between $B$ and $C$ are: $B$ is independent of $C$. $B$ and $C$ are independent given $A$. Since the middle variable $D$ is not in any conditioning set that makes $B$ and $C$ independent $\Rightarrow$ $(B,D,C)$ form a v-structure $\Rightarrow$ we orient the arrows towards the middle node $D$: $B \rightarrow D$, $C \rightarrow D$. \\
        In total: $A \rightarrow D$, $C \rightarrow D$, $B \rightarrow D$, $C \rightarrow D$, so removing duplicates: $A \rightarrow D$, $C \rightarrow D$, $B \rightarrow D$. \\
        \textbf{Answer}: $A \rightarrow D$, $C \rightarrow D$, $B \rightarrow D$ \\
        \vspace{1mm}
        \textbf{Question}: Premise: Suppose there is a closed system of 4 variables, $A$, $B$, $C$ and $D$. All the statistical relations among these 4 variables are as follows: $A$ correlates with $B$. $A$ correlates with $C$. $A$ correlates with $D$. $B$ correlates with $C$. $B$ correlates with $D$. $C$ correlates with $D$. However, $A$ is independent of $C$ given $B$. $B$ is independent of $D$ given $C$. $A$ is independent of $D$ given $B$. $A$ is independent of $D$ given $C$. $A$ is independent of $D$ given $B$ and $C$.
        Given the possible v-structures from the previous step: $(A,B,C)$, $(B,C,D)$. 
        Can you apply the second step of the \pc algorithm to orient these v-structures? \\
        \vspace{-1.4mm}
        \textbf{Reasoning}: We first go through all possible v-structures: \\
        $(A,B,C)$ is a possible v-structure: all conditional independences between $A$ and $C$ are: $A$ is independent of $C$ given $B$. Since the middle variable $B$ is in a conditioning set that makes $A$ and $C$ independent $\Rightarrow$ $(A,B,C)$ does not form a v-structure $\Rightarrow$ No directed edges found. \\
        $(B,C,D)$ is a possible v-structure: all conditional independences between $B$ and $D$ are: $B$ is independent of $D$ given $C$. Since the middle variable $C$ is in a conditioning set that makes $B$ and $D$ independent $\Rightarrow$ $(B,C,D)$ does not form a v-structure $\Rightarrow$ No directed edges found. \\
        \vspace{-1.4mm}
        \textbf{Answer}: No directed edges found \\
        \vspace{1mm}
        \textbf{Question}: Premise: Suppose there is a closed system of 3 variables, $A$, $B$ and $C$. All the statistical relations among these 3 variables are as follows: $A$ correlates with $C$. $B$ correlates with $C$. However, $A$ is independent of $B$.
        Given the possible v-structures from the previous step: $(A,C,B)$.
        Can you apply the second step of the \pc algorithm to orient these v-structures? \\
        \vspace{-1.4mm}
        \textbf{Reasoning}: We first go through all possible v-structures: \\
        $(A,C,B)$ is a possible v-structure: all conditional independences between $A$ and $B$ are: $A$ is independent of $B$. Since the middle variable $C$ is not in any conditioning set that makes $A$ and $B$ independent $\Rightarrow$ $(A,C,B)$ form a v-structure $\Rightarrow$ we orient the arrows towards the middle node $C$: $A \rightarrow C$, $B \rightarrow C$. \\
        \vspace{-1.4mm}
        \textbf{Answer}: $A \rightarrow C$, $B \rightarrow C$ \\
        \vspace{1mm}
        \textbf{Question}: \texttt{Premise}
        Given the possible v-structures from the previous step: \texttt{[Answer to SubQ4]} \\
        Can you apply the second step of the PC algorithm to orient these v-structures? \\
        \vspace{-1.4mm}
        \textbf{Reasoning}: \\ 
        \bottomrule
    \end{tabular}
    \label{tab:appendix-subq5}
\end{table}

\begin{table}[h!]
    \caption{    
    Few-shot exemplars for \subq6 followed by \subq6 at the end.
    }
    \centering
    \small
    \begin{tabular}{p{\linewidth}}
        \toprule
        \textbf{Question}: Can you orient the edges of the undirected graph: $(A,B), (B,C), (B,D), (C,E), (D,E)$, according to the directed edges: $C \rightarrow E, D \rightarrow E$? \\
        \vspace{-1.4mm}
        \textbf{Reasoning}: \\
        $(A,B)$ remains $(A,B)$ \\
        $(B,C)$ remains $(B,C)$ \\
        $(B,D)$ remains $(B,D)$ \\
        $(C,E)$ becomes $C \rightarrow E$ \\
        $(D,E)$ becomes $D \rightarrow E$. \\
        So the final (partially) directed graph is: $(A,B), (B,C), (B,D), C \rightarrow E, D \rightarrow E$. \\
        \vspace{-1.4mm}
        \textbf{Answer}: $(A,B), (B,C), (B,D), C \rightarrow E, D \rightarrow E$ \\
        \vspace{1mm}
        \textbf{Question}: Can you orient the edges of the undirected graph: $(A,B), (B,C), (A,C)$, according to the directed edges: No directed edges found? \\
        \vspace{-1.4mm}
        \textbf{Reasoning}: Since there are no directed edges $\Rightarrow$ the final (undirected) graph remains: $(A,B), (B,C), (A,C)$. \\
        \vspace{-1.4mm}
        \textbf{Answer}: $(A,B), (B,C), (A,C)$ \\
        \vspace{1mm}
        \textbf{Question}: Can you orient the edges of the undirected graph: \texttt{[Answer to SubQ2]}, according to the directed edges: \texttt{[Answer to SubQ5]} \\
        \vspace{-1.4mm}
        \textbf{Reasoning}: \\ 
        \bottomrule
    \end{tabular}
    \label{tab:appendix-subq6}
\end{table}

\begin{table}[h!]
    \caption{    
    Few-shot exemplars for \subq7 followed by \subq7 at the end.
    }
    \centering
    \small
    \begin{tabular}{p{\linewidth}}
        \toprule
        \textbf{Question}: Given the (partially) directed graph of the second step of the PC algorithm: $(A,B), (A,C), (B,C)$ and all paths of length 2 in the graph: $(B,A,C), (A,B,C), (A,C,B)$.
        Can you apply the third step of the PC algorithm to possibly orient remaining edges of the partially directed graph? \\
        \vspace{-1.4mm}
        \textbf{Reasoning}: We go through all paths of length 2 in the graph: $(B,A,C), (A,B,C), (A,C,B)$ and possibly orient non directed edges. In our case: \\
        For path $(B,A,C)$ with middle node $A$: Both $(A,B)$ and $(A,C)$ are undirected $\Rightarrow$ We do not orient an edge. \\
        For path $(A,B,C)$ with middle node $B$: Both $(A,B)$ and $(B,C)$ are undirected $\Rightarrow$ We do not orient an edge. \\
        For path $(A,C,B)$ with middle node $C$: Both $(A,C)$ and $(B,C)$ are undirected $\Rightarrow$ We do not orient an edge. \\
        So the final graph is: $(A,B), (A,C), (B,C)$. \\
        \vspace{-1.4mm}
        \textbf{Answer}: $(A,B), (A,C), (B,C)$ \\
        \vspace{1mm}
        \textbf{Question}: Given the (partially) directed graph of the second step of the PC algorithm: $A \rightarrow C, B \rightarrow C, (C,D)$ and all paths of length 2 in the graph: $(A,C,B), (A,C,D), (B,C,D)$. \\
        Can you apply the third step of the PC algorithm to possibly orient remaining edges of the partially directed graph? \\
        \vspace{-1.4mm}
        \textbf{Reasoning}: We go through all paths of length 2 in the graph: $(A,C,B), (A,C,D), (B,C,D)$ and possibly orient non directed edges. In our case: \\
        For path $(A,C,B)$ with middle node $C$: Both $A \rightarrow C$ and $B \rightarrow C$ are directed $\Rightarrow$ No undirected edge to orient. \\
        For path $(A,C,D)$ with middle node $C$: $A \rightarrow C$ is directed towards the middle node $C$ and $(C,D)$ is undirected $\Rightarrow$ We orient $C \rightarrow D$ such that no extra v-structure is created. \\
        For path $(B,C,D)$ with middle node $C$: $B \rightarrow C$ is directed towards the middle node $C$ and $(C,D)$ is undirected  $\Rightarrow$  we orient $C \rightarrow D$ such that no extra v-structure is created. \\
        So, the final graph is: $A \rightarrow C, B \rightarrow C, C \rightarrow D$. \\
        \vspace{-1.4mm}
        \textbf{Answer}: $A \rightarrow C, B \rightarrow C, C \rightarrow D$ \\
        \vspace{1mm}
        \textbf{Question}: Given the (partially) directed graph of the second step of the PC algorithm: $A \rightarrow C, B \rightarrow C, (A,D)$ and all paths of length 2 in the graph: $(A,C,B), (C,A,D)$. \\
        Can you apply the third step of the PC algorithm to possibly orient remaining edges of the partially directed graph? \\
        \vspace{-1.4mm}
        \textbf{Reasoning}: We go through all paths of length 2 in the graph: $(A,C,B), (C,A,D)$ and possibly orient the non directed edges. In our case: \\
        For path $(A,C,B)$ with middle node $C$: Both $A \rightarrow C$ and $B \rightarrow C$ are directed $\Rightarrow$ No undirected edge to orient. \\
        For path $(C,A,D)$ with middle node $A$: $A \rightarrow C$ is directed towards the outer node $C$ and $(A,D)$ is undirected $\Rightarrow$ We do not orient an edge.
        So, the final graph is: $A \rightarrow C, B \rightarrow C, (A,D)$. \\
        \vspace{-1.4mm}
        \textbf{Answer}: $A \rightarrow C, B \rightarrow C, (A,D)$ \\
        \vspace{1mm}
        \textbf{Question}: Given the (partially) directed graph of the second step of the PC algorithm: \texttt{[Answer to SubQ6]} and all paths of length 2 in the graph: \texttt{[Answer to SubQ3]}
        Can you apply the third step of the PC algorithm to possibly orient remaining edges of the partially directed graph? \\
        \vspace{-1.4mm}
        \textbf{Reasoning}: \\ 
        \bottomrule
    \end{tabular}
    \label{tab:appendix-subq7}
\end{table}

\begin{table}[h!]
    \caption{    
    \nth{1}-\nth{5} exemplars (out of 11 shots in total) for \subq8.
    }
    \centering
    \small
    \begin{tabular}{p{\linewidth}}
        \toprule
        \textbf{Question}: Given the causal graph: $A \rightarrow C, B \rightarrow C$.
        Can you infer if the Hypothesis "$B$ directly affects $C$." is True or False? \\
        \vspace{-1.4mm}
        \textbf{Reasoning}: We have $B$ directly affects $C$, that is $B \rightarrow C$.
        The directed edge $B \rightarrow C$ belongs to the set $\{A \rightarrow C, B \rightarrow C\}$, so the hypothesis is True. \\
        \vspace{-1.4mm}
        \textbf{Answer}: 1 \\
        \vspace{1mm}
        \textbf{Question}: Given the causal graph: $A \rightarrow C, B \rightarrow C$. 
        Can you infer if the Hypothesis "$A$ directly affects $B$." is True or False? \\
        \vspace{-1.4mm}
        \textbf{Reasoning}: We have $A$ directly affects $B$, that is $A \rightarrow B$.
        The directed edge $A \rightarrow B$ does not belong to the set $\{A \rightarrow C, B \rightarrow C\}$, so the hypothesis is False. \\
        \vspace{-1.4mm}
        \textbf{Answer}: 0 \\
        \vspace{1mm}
        \textbf{Question}: Given the causal graph: $(C,D)$.
        Can you infer if the Hypothesis "$C$ directly affects $D$." is True or False? \\
        \vspace{-1.4mm}
        \textbf{Reasoning}: We have $C$ directly affects $D$, that is $C \rightarrow D$.
        The directed edge $C \rightarrow D$ does not belong to the set $\{(C,D)\}$ (since $C \rightarrow D$ is directed while $(C,D)$ is undirected), so the hypothesis is False. \\
        \vspace{-1.4mm}
        \textbf{Answer}: 0 \\
        \vspace{1mm}
        \textbf{Question}: Given the causal graph: $(A,B), (B,C), (B,D), C \rightarrow E, D \rightarrow E$. 
        Can you infer if the Hypothesis "$B$ and $C$ together cause some other variable(s)." is True or False? \\
        \vspace{-1.4mm}
        \textbf{Reasoning}: We have $B$ and $C$ together cause some other variable(s).
        We go through all variables apart from $B$ and $C$: $A, D, E$ and check if there is a directed edge from $B$ to it and another one from $C$ to it: \\
        For $A$: the directed edge $B \rightarrow A$ does not belong to the set $\{(A,B)$, $(B,C)$, $(B,D)$, $C \rightarrow E$, $D \rightarrow E\}$ and the directed edge $C \rightarrow A$ does not belong to the set $\{(A,B)$, $(B,C)$, $(B,D)$, $C \rightarrow E$, $D \rightarrow E\}$. \\
        For $D$: the directed edge $B \rightarrow D$ does not belong to the set $\{(A,B)$, $(B,C)$, $(B,D)$, $C \rightarrow E$, $D \rightarrow E\}$ (since $B \rightarrow D$ is directed while $(B,D)$ is undirected) and the directed edge $C \rightarrow D$ does not belong to the set $\{(A,B)$, $(B,C)$, $(B,D)$, $C \rightarrow E$, $D \rightarrow E\}$. \\
        For $E$: the directed edge $B \rightarrow E$ does not belong to the set $\{(A,B)$, $(B,C)$, $(B,D)$, $C \rightarrow E$, $D \rightarrow E\}$ and the directed edge $C \rightarrow E$ belongs to the set $\{(A,B)$, $(B,C), (B,D)$, $C \rightarrow E$, $D \rightarrow E\}$. \\
        From the above, there is no variable in the causal graph $\{(A,B)$, $(B,C)$, $(B,D)$, $C \rightarrow E$, $D \rightarrow E\}$ with an incoming edge both from $B$ and another one from $C$, so the hypothesis is False. \\
        \vspace{-1.4mm}
        \textbf{Answer}: 0 \\
        \vspace{1mm}
        \textbf{Question}: Given the causal graph: $(A,B), (B,C), (B,D), C \rightarrow E, D \rightarrow E$.
        Can you infer if the Hypothesis "$A$ and $B$ together cause some other variable(s)." is True or False? \\
        \vspace{-1.4mm}
        \textbf{Reasoning}: We have $A$ and $E$ together cause some other variable(s).
        We go through all variables apart from $A$ and $E$: $B, C, D$ and check if there is a directed edge from $A$ to it and another one from $E$ to it: \\
        For $B$: the directed edge $A \rightarrow B$ does not belong to the set $\{(A,B)$, $(B,C)$, $(B,D)$, $C \rightarrow E$, $D \rightarrow E\}$ (since $A \rightarrow B$ is directed while $(A,B)$ is undirected) and the directed edge $E \rightarrow B$ does not belong to the set $\{(A,B)$, $(B,C)$, $(B,D)$, $C \rightarrow E$, $D \rightarrow E\}$. \\
        For $C$: the directed edge $A \rightarrow C$ does not belong to the set $\{(A,B)$, $(B,C)$, $(B,D)$, $C \rightarrow E$, $D \rightarrow E\}$ and the directed edge $E \rightarrow C$ does not belong to the set $\{(A,B)$, $(B,C)$, $(B,D)$, $C \rightarrow E$, $D \rightarrow E\}$. \\
        For $D$: the directed edge $A \rightarrow D$ does not belong to the set $\{(A,B)$, $(B,C)$, $(B,D)$, $C \rightarrow E$, $D \rightarrow E\}$ and the directed edge $E \rightarrow D$ does not belong to the set $\{(A,B)$, $(B,C)$, $(B,D)$, $C \rightarrow E$, $D \rightarrow E\}$. \\
        From the above, there is no variable in the causal graph $\{(A,B)$, $(B,C)$, $(B,D)$, $C \rightarrow E$, $D \rightarrow E\}$ with an incoming edge both from $A$ and another one from $E$, so the hypothesis is False. \\
        \vspace{-1.4mm}
        \textbf{Answer}: 0 \\
        \bottomrule
    \end{tabular}
    \label{tab:appendix-subq8-a}
\end{table}

\begin{table}[h!]
    \caption{    
    \nth{6}-\nth{9} exemplars (out of 11 shots in total) for \subq8.
    }
    \centering
    \small
    \begin{tabular}{p{\linewidth}}
        \toprule
        \textbf{Question}: Given the causal graph: $(A,B), (B,C), (B,D), C \rightarrow E, D \rightarrow E$. \\
        Can you infer if the Hypothesis "Some variable(s) cause both $C$ and $D$." is True or False? \\
        \vspace{-1.4mm}
        \textbf{Reasoning}: We have Some variable(s) cause both $C$ and $D$.
        We go through all variables apart from $C$ and $D$: $A, B, E$ and check if there is a directed edge from it to $C$ and another one from it to $D$: \\
        For $A$: the directed edge $A \rightarrow C$ does not belong to the set $\{(A,B)$, $(B,C)$, $(B,D)$, $C \rightarrow E$, $D \rightarrow E\}$ and the directed edge $A \rightarrow D$ does not belong to the set $\{(A,B)$, $(B,C)$, $(B,D)$, $C \rightarrow E$, $D \rightarrow E\}$. \\
        For $B$: the directed edge $B \rightarrow C$ does not belong to the set $\{(A,B)$, $(B,C)$, $(B,D)$, $C \rightarrow E$, $D \rightarrow E\}$ (since $B \rightarrow C$ is directed while $(B,C)$ is undirected) and the directed edge $B \rightarrow D$ does not belong to the set $\{(A,B)$, $(B,C)$, $(B,D)$, $C \rightarrow E$, $D \rightarrow E\}$ (since $B \rightarrow D$ is directed while $(B,D)$ is undirected). \\
        For $E$: the directed edge $E \rightarrow C$ does not belong to the set $\{(A,B)$, $(B,C)$, $(B,D)$, $C \rightarrow E$, $D \rightarrow E\}$ and the directed edge $E \rightarrow D$ does not belong to the set $\{(A,B)$, $(B,C)$, $(B,D)$, $C \rightarrow E$, $D \rightarrow E\}$. \\
        From the above, there is no variable in the causal graph $\{(A,B)$, $(B,C)$, $(B,D)$, $C \rightarrow E$, $D \rightarrow E\}$ with an outgoing edge pointing to $C$ and another one pointing to $D$, so the hypothesis is False. \\
        \vspace{-1.4mm}
        \textbf{Answer}: 0 \\
        \vspace{1mm}
        \textbf{Question}: Given the causal graph: $A \rightarrow C, B \rightarrow C, C \rightarrow D$.
        Can you infer if the Hypothesis "$A$ influences $D$ through some mediator(s)." is True or False? \\
        \vspace{-1.4mm}
        \textbf{Reasoning}: We have $A$ influences $D$ through some mediator(s).
        We go through all variables apart from $A$ and $D$: $B$, $C$ and check if there is a directed edge from $A$ to it and another one from it to $D$ (note we only care about directed edges not undirected ones): \\
        For $B$: we have edges $A \rightarrow B$ and $B \rightarrow D$. The directed edge $A \rightarrow B$ does not belong to the set $\{A \rightarrow C, B \rightarrow C, C \rightarrow D\}$ and the directed edge $B \rightarrow D$ does not belong to the set $\{A \rightarrow C, B \rightarrow C, C \rightarrow D\}$. Since none of the edges belong to the graph, so $B$ is not a mediator. \\
        For $C$: we have edges $A \rightarrow C$ and $C \rightarrow D$. Both the directed edge $A \rightarrow C$ belongs to the set $\{A \rightarrow C, B \rightarrow C, C \rightarrow D\}$ and the directed edge $C \rightarrow D$ belongs to the set $\{A \rightarrow C, B \rightarrow C, C \rightarrow D\}$, so $C$ is a mediator. \\
        From the above, both $A \rightarrow C$ and $C \rightarrow D$ belong the set of edges $\{A \rightarrow C, B \rightarrow C, C \rightarrow D\}$, so the hypothesis is True. \\
        \vspace{-1.4mm}
        \textbf{Answer}: 1 \\
        \vspace{1mm}
        \textbf{Question}: Given the causal graph: $(A,B), (B,C), (B,D), C \rightarrow E, D \rightarrow E$.
        Can you infer if the Hypothesis "$B$ influences $E$ through some mediator(s)." is True or False? \\
        \vspace{-1.4mm}
        \textbf{Reasoning}: We have $B$ influences $E$ through some mediator(s).
        We go through all variables apart from $B$ and $E$: $A$, $C$, $D$ and check if there is a directed edge from $B$ to it and another one from it to $E$: \\
        For $A$: we have edges $B \rightarrow A$ and $A \rightarrow E$. The directed edge $B \rightarrow A$ does not belong to the set $\{(A,B), (B,C), (B,D), C \rightarrow E, D \rightarrow E\}$ and the directed edge $A \rightarrow E$ does not belong to the set $\{(A,B), (B,C), (B,D), C \rightarrow E, D \rightarrow E\}$. Since none of the edges belong to the graph, so $A$ is not a mediator. \\
        For $C$: we have edges $B \rightarrow C$ and $C \rightarrow E$. The directed edge $B \rightarrow C$ does not belong to the set $\{(A,B), (B,C), (B,D), C \rightarrow E, D \rightarrow E\}$ (since $B \rightarrow C$ is directed while $(B,C)$ is undirected) but the directed edge $C \rightarrow E$ belongs to $\{(A,B), (B,C), (B,D), C \rightarrow E, D \rightarrow E\}$. Since only one edge, $C \rightarrow E$, belongs to the set of edges, so $C$ is not a mediator. \\
        For $D$: we have edges $B \rightarrow D$ and $D \rightarrow E$. The directed edge $B \rightarrow D$ does not belong to the set $\{(A,B), (B,C), (B,D), C \rightarrow E, D \rightarrow E\}$ (since $B \rightarrow D$ is directed while $(B,D)$ is undirected) but the directed edge $D \rightarrow E$ belongs to $\{(A,B), (B,C), (B,D), C \rightarrow E, D \rightarrow E\}$. Since only one edge, $D \rightarrow E$, belongs to the set of edges, so $D$ is not a mediator. \\
        From the above, $B$ does not influence $E$ through some mediator(s), so the hypothesis is False. \\
        \vspace{-1.4mm}
        \textbf{Answer}: 0 \\
        \vspace{1mm}
        \textbf{Question}: Given the inferred causal graph: $A \rightarrow C, B \rightarrow C, C \rightarrow D$.
        Can you infer if the Hypothesis "$A$ influences $B$ through some mediator(s)." is True or False? \\
        \vspace{-1.4mm}
        \textbf{Reasoning}: We have $A$ influences $B$ through some mediator(s).
        We go through all variables apart from $A$ and $B$: $C$, $D$ and check if there is a directed edge from $A$ to it and another one from it to $B$: \\
        For $C$: we have edges $A \rightarrow C$ and $C \rightarrow B$. The directed edge $A \rightarrow C$ belongs to the set of edges $\{A \rightarrow C, B \rightarrow C, C \rightarrow D\}$ but the directed edge $C \rightarrow B$ does not belong to the set of edges $\{A \rightarrow C, B \rightarrow C, C \rightarrow D\}$. Since only one edge, $A \rightarrow C$, belongs to the set of edges, so $C$ is not a mediator. \\
        For $D$: we have edges $A \rightarrow D$ and $D \rightarrow B$. The directed edge $A \rightarrow D$ does not belong to the set of edges $\{A \rightarrow C, B \rightarrow C, C \rightarrow D\}$ and the directed edge $D \rightarrow B$ does not belong to the set of edges $\{A \rightarrow C, B \rightarrow C, C \rightarrow D\}$. Since none of the edges belongs to the graph, so $D$ is not a mediator. \\
        From the above, $A$ does not influence $B$ through some mediator(s), so the hypothesis is False. \\
        \vspace{-1.4mm}
        \textbf{Answer}: 0 \\
        \bottomrule
    \end{tabular}
    \label{tab:appendix-subq8-b}
\end{table}

\begin{table}[h!]
    \caption{    
    \nth{10} and \nth{11} exemplars (out of 11 shots in total) for \subq8 followed by \subq8 at the end.
    }
    \centering
    \small
    \begin{tabular}{p{\linewidth}}
        \toprule
        \textbf{Question}: Given the causal graph: $A \rightarrow C, B \rightarrow C, C \rightarrow D$.
        Can you infer if the Hypothesis "$C$ influences $D$ through some mediator(s)." is True or False? \\
        \vspace{-1.4mm}
        \textbf{Reasoning}: We have $C$ influences $D$ through some mediator(s).
        We go through all variables apart from $C$ and $D$: $A$, $B$ and check if there is a directed edge from $C$ to it and another one from it to $D$: \\
        For $A$: we have edges $C \rightarrow A$ and $A \rightarrow D$. The directed edge $C \rightarrow A$ does not belong to the set $\{A \rightarrow C, B \rightarrow C, C \rightarrow D\}$ and the directed edge $A \rightarrow D$ does not belong to the set $\{A \rightarrow C, B \rightarrow C, C \rightarrow D\}$. Since none of the edges belongs to the graph, so $A$ is not a mediator. \\
        For $B$: we have edges $C \rightarrow B$ and $B \rightarrow D$. The directed edge $C \rightarrow B$ does not belong to the set $\{A \rightarrow C, B \rightarrow C, C \rightarrow D\}$ and the directed edge $B \rightarrow D$ does not belong to the set $\{A \rightarrow C, B \rightarrow C, C \rightarrow D\}$. Since none of the edges belongs to the graph, so $B$ is not a mediator. \\
        From the above, $C$ does not influence $D$ through some mediator(s), so the hypothesis is False. \\
        \vspace{-1.4mm}
        \textbf{Answer}: 0 \\
        \vspace{1mm}
        \textbf{Question}: Given the causal graph: $(A,B), (B,C)$.
        Can you infer if the Hypothesis "$A$ directly affects $B$." is True or False? \\
        \vspace{-1.4mm}
        \textbf{Reasoning}: We have $A$ directly affects $B$, that is $A \rightarrow B$.
        The directed edge $A \rightarrow B$ does not belong to the set $\{(A,B), (B,C)\}$ (since $A \rightarrow B$ is directed while $(A,B)$ is undirected), so the hypothesis is False. \\
        \vspace{-1.4mm}
        \textbf{Answer}: 0 \\
        \vspace{1mm}
        \textbf{Question}: Given the inferred final causal graph: \texttt{[Answer to SubQ7]}
        Can you infer if the Hypothesis \texttt{[Hypothesis]} is True or False? \\
        Keep in mind that directed edges, for example $A \rightarrow B$, are different from undirected edges, for example $(A,B)$. \\
        \vspace{-1.4mm}
        \textbf{Reasoning}: \\ 
        \bottomrule
    \end{tabular}
    \label{tab:appendix-subq8-c}
\end{table}

\clearpage

\section{Baseline Prompting Strategies: Few-shot and Few-shot \cotp Examples}\label{sec:appendix_baseline_shots}

Tables \ref{tab:appendix-baselines-a} and \ref{tab:appendix-baselines-b} provide the 6-shot \cotp examples for the corresponding baseline strategy in Fig. \ref{fig:baselines}. The exact same examples, excluding the \textit{Reasoning} part, were used for the few-shot baseline strategy. 
For PaLM 2 L, the Reasoning parts of the last three shots were shortened to fit the context limit.

\begin{table}[h!]
    \caption{  
    \nth{1}-\nth{4} exemplars (out of 6 shots in total) for few-shot \cotp prompting strategy. The exact same exemplars are used for few-shot strategy but without the "Reasoning" part.
    }
    \centering
    \small
    \begin{tabular}{p{\linewidth}}
        \toprule
        \textbf{Question}: Premise: Suppose there is a closed system of 3 variables, $A, B$ and $C$. All the statistical relations among these 3 variables are as follows: $A$ correlates with $B$. $A$ correlates with $C$. $B$ correlates with $C$. Hypothesis: $B$ directly affects $C$.
        Is the Hypothesis True or False? Answer with 0 or 1. \\
        \vspace{-1.4mm}
        \textbf{Reasoning}: The initial fully connected undirected graph is $(A,B), (A,C), (B,C)$. Since we have no conditional independencies we do not remove any edges. We have $B$ directly affects $C$, that is $B \rightarrow C$. The directed edge $B \rightarrow C$ does not belong to the set $\{(A,B), (A,C), (B,C)\}$, so the hypothesis is False. \\
        \vspace{-1.4mm}
        \textbf{Answer}: 0 \\
        \vspace{1mm}
        \textbf{Question}: Premise: Suppose there is a closed system of 3 variables, $A, B$ and $C$. All the statistical relations among these 3 variables are as follows: $A$ correlates with $B$. However, $A$ is independent of $C$. $A$ and $C$ are independent given $B$. $B$ is independent of $C$. $B$ and $C$ are independent given $A$. Hypothesis: $A$ influences $B$ through some mediator(s).
        Is the Hypothesis True or False? Answer with 0 or 1. \\
        \vspace{-1.4mm}
        \textbf{Reasoning}: The initial fully connected undirected graph is $(A,B), (A,C), (B,C)$. Since $A$ is independent of $B \Rightarrow (A,B)$ is removed $\Rightarrow$ the graph after this step is $(A,C), (B,C)$. Since $B$ is independent of $C \Rightarrow (B,C)$ is removed $\Rightarrow$ the graph after this step is $(A,C)$. So, $A$ does not influence $B$ through some mediator(s), so the hypothesis is False. \\
        \vspace{-1.4mm}
        \textbf{Answer}: 0 \\
        \vspace{1mm}
        \textbf{Question}: Premise: Suppose there is a closed system of 3 variables, $A, B$ and $C$. All the statistical relations among these 3 variables are as follows: $A$ correlates with $B$. However, $A$ is independent of $C$. $A$ and $C$ are independent given $B$. $B$ is independent of $C$. $B$ and $C$ are independent given $A$. Hypothesis: Some variable(s) cause(s) both $C$ and $B$. 
        Is the Hypothesis True or False? Answer with 0 or 1. \\
        \vspace{-1.4mm}
        \textbf{Reasoning}: The initial fully connected undirected graph is $(A,B), (A,C), (B,C)$. Since $A$ is independent of $C \Rightarrow (A,C)$ is removed $\Rightarrow$ the graph after this step is $(A,B), (B,C)$. Since $B$ is independent of $C \Rightarrow (B,C)$ is removed $\Rightarrow$ the graph after this step is $(A,B)$. So, there is no variable causing both C and B, so the hypothesis is False. \\
        \vspace{-1.4mm}
        \textbf{Answer}: 0 \\
        \vspace{1mm}
        \textbf{Question}: Premise: Suppose there is a closed system of 4 variables, $A, B, C$ and $D$. All the statistical relations among these 4 variables are as follows: $A$ correlates with $B$. $A$ correlates with $C$. $A$ correlates with $D$. $B$ correlates with $C$. $B$ correlates with $D$. $C$ correlates with $D$. However, $B$ and $C$ are independent given $A$. Hypothesis: $D$ and $A$ together cause some other variable(s).
        Is the Hypothesis True or False? Answer with 0 or 1. \\
        \vspace{-1.4mm}
        \textbf{Reasoning}: The initial fully connected undirected graph is $(A,B), (A,C), (A,D), (B,C), (B,D), (C,D)$. Since $B$ and $C$ are independent given $A \Rightarrow (B,C)$ is removed $\Rightarrow$ the graph after this step is $(A,B), (A,C), (A,D), (B,D), (C,D)$. For path $(C,A,B)$: there is no edge $(C,B)$ but $B$ and $C$ are independent given $A$, so $(C,A,B)$ is not a v-structure. For path $(C,D,B)$: there is no edge $(C,B)$ and $B$ and $C$ are not independent given $D$, so $(C,D,B)$ is a v-structure. So the final graph is: $(A,B), (A,C), (A,D), (B,C), B \rightarrow D, C \rightarrow D$. So, $D$ and $A$ together do not cause some other variable(s), so the hypothesis is False. \\
        \vspace{-1.4mm}
        \textbf{Answer}: 0 \\
        \bottomrule
    \end{tabular}
    \label{tab:appendix-baselines-a}
\end{table}

\begin{table}[h!]
    \caption{    
    \nth{5}-\nth{6} exemplars (out of 6 shots in total) for few-shot \cotp prompting strategy. The exact same exemplars are used for few-shot strategy but without the "Reasoning" part.
    }
    \centering
    \small
    \begin{tabular}{p{\linewidth}}
        \toprule
        \textbf{Question}: Premise: Suppose there is a closed system of 4 variables, $A, B, C$ and $D$. All the statistical relations among these 4 variables are as follows: $A$ correlates with $B$. $A$ correlates with $C$. $A$ correlates with $D$. $B$ correlates with $C$. $B$ correlates with $D$. $C$ correlates with $D$. However, $B$ and $C$ are independent given $A$. Hypothesis: $A$ directly affects $B$. 
        Is the Hypothesis True or False? Answer with 0 or 1. \\
        \vspace{-1.4mm}
        \textbf{Reasoning}: The initial fully connected undirected graph is $(A,B)$, $(A,C)$, $(A,D)$, $(B,C)$, $(B,D)$, $(C,D)$. Since $B$ and $C$ are independent given $A$ $\Rightarrow$ $(B,C)$ is removed $\Rightarrow$ the graph after this step is $(A,B), (A,C), (A,D), (B,D), (C,D)$. For path $(C,A,B)$: there is no edge $(C,B)$ but $B$ and $C$ are independent given $A$, so $(C,A,B)$ is not a v-structure. For path $(C,D,B)$: there is no edge $(C,B)$ and $B$ and $C$ are not independent given $D$, so $(C,D,B)$ is a v-structure. So the final graph is: $(A,B), (A,C), (A,D), (B,C), B \rightarrow D, C \rightarrow D$. We have $A$ directly affects $B$, that is $A \rightarrow B$. The directed edge $A \rightarrow B$ does not belong to the set $\{(A,B), (A,C), (A,D), (B,C), B \rightarrow D, C \rightarrow D\}$, so the hypothesis is False. \\
        \vspace{-1.4mm}
        \textbf{Answer}: 0 \\
                \vspace{1mm}
        \textbf{Question}: Premise: Suppose there is a closed system of 4 variables, $A, B, C$ and $D$. All the statistical relations among these 4 variables are as follows: $A$ correlates with $B$. $A$ correlates with $C$. $A$ correlates with $D$. $B$ correlates with $C$. $B$ correlates with $D$. $C$ correlates with $D$. However, $B$ and $C$ are independent given $A$. Hypothesis: $B$ and $C$ together cause some other variable(s).
        Is the Hypothesis True or False? Answer with 0 or 1. \\
        \vspace{-1.4mm}
        \textbf{Reasoning}: The initial fully connected undirected graph is $(A,B)$, $(A,C)$, $(A,D)$, $(B,C)$, $(B,D)$, $(C,D)$. Since $B$ and $C$ are independent given $A$ $\Rightarrow$ $(B,C)$ is removed $\Rightarrow$ the graph after this step is $(A,B), (A,C), (A,D), (B,D), (C,D)$. For path $(C,A,B)$: there is no edge $(C,B)$ but $B$ and $C$ are independent given $A$, so $(C,A,B)$ is not a v-structure. For path $(C,D,B)$: there is no edge $(C,B)$ and $B$ and $C$ are not independent given $D$, so $(C,D,B)$ is a v-structure. So the final graph is: $(A,B), (A,C), (A,D), (B,C), B \rightarrow D, C \rightarrow D$. So, $B$ and $C$ together cause some other variable, namely $D$, so the hypothesis is True. \\
        \vspace{-1.4mm}
        \textbf{Answer}: 1 \\
        \vspace{1mm}
        \textbf{Question}: \texttt{[Premise]} \texttt{[Hypothesis]}
        Is the Hypothesis True or False? Answer with 0 or 1. \\
        \vspace{-1.4mm}
        \textbf{Reasoning}: \\ 
        \bottomrule
    \end{tabular}
    \label{tab:appendix-baselines-b}
\end{table}

\clearpage

\section{\ourprompting: Refactoring and Paraphrasing}\label{sec:appendix_perturbing}

Table \ref{tab:appendix-pertubing} shows an example of how we refactor (rename) and paraphrase \ourprompting prompts based on the robustness tests of \citet{jin2024can}. For more details, see the robustness analysis in Sec. \ref{sec:results}.
\begin{table}[h!]
    \caption{    
    Example of refactoring and paraphrasing the \nth{5} shot of \subq8.
    }
    \centering
    \small
    \begin{tabular}{p{\linewidth}}
        \toprule
        \underline{\textbf{\textsc{Original}}} \\
        \vspace{-0.5mm}
        \textbf{Question}: Given the causal graph: $(A,B), (B,C), (B,D), C \rightarrow E, D \rightarrow E$.
        Can you infer if the Hypothesis "$A$ and $B$ together cause some other variable(s)." is True or False? \\
        \vspace{-1.4mm}
        \textbf{Reasoning}: We have $A$ and $E$ together cause some other variable(s).
        We go through all variables apart from $A$ and $E$: $B, C, D$ and check if there is a directed edge from $A$ to it and another one from $E$ to it: \\
        For $B$: the directed edge $A \rightarrow B$ does not belong to the set $\{(A,B)$, $(B,C)$, $(B,D)$, $C \rightarrow E$, $D \rightarrow E\}$ (since $A \rightarrow B$ is directed while $(A,B)$ is undirected) and the directed edge $E \rightarrow B$ does not belong to the set $\{(A,B)$, $(B,C)$, $(B,D)$, $C \rightarrow E$, $D \rightarrow E\}$. \\
        For $C$: the directed edge $A \rightarrow C$ does not belong to the set $\{(A,B)$, $(B,C)$, $(B,D)$, $C \rightarrow E$, $D \rightarrow E\}$ and the directed edge $E \rightarrow C$ does not belong to the set $\{(A,B)$, $(B,C)$, $(B,D)$, $C \rightarrow E$, $D \rightarrow E\}$. \\
        For $D$: the directed edge $A \rightarrow D$ does not belong to the set $\{(A,B)$, $(B,C)$, $(B,D)$, $C \rightarrow E$, $D \rightarrow E\}$ and the directed edge $E \rightarrow D$ does not belong to the set $\{(A,B)$, $(B,C)$, $(B,D)$, $C \rightarrow E$, $D \rightarrow E\}$. \\
        From the above, there is no variable in the causal graph $\{(A,B)$, $(B,C)$, $(B,D)$, $C \rightarrow E$, $D \rightarrow E\}$ with an incoming edge both from $A$ and another one from $E$, so the hypothesis is False. \\
        \textbf{Answer}: 0 \\
        \hline \\
        \vspace{-3mm}
        \underline{\textbf{\textsc{Refactored}}} \\
        \vspace{-0.5mm}
        \textbf{Question}: Given the causal graph: $(Z,Y), (Y,X), (Y,W), X \rightarrow V, W \rightarrow V$. Can you infer if the Hypothesis "Z and Y together cause some other variable(s)." is True or False? \\
        \vspace{-1.4mm}
        \textbf{Reasoning}: We have $Z$ and $V$ together cause some other variable(s).
        We go through all variables apart from $Z$ and $V$: $Y$, $X$, $W$ and check if there is a directed edge from $Z$ to it and another one from $V$ to it: \\
        For $Y$: the directed edge $Z \rightarrow Y$ does not belong to the set $\{(Z,Y)$, $(Y,X)$, $(Y,W)$, $X \rightarrow V$, $W \rightarrow V\}$ (since $Z \rightarrow Y$ is directed while $(Z,Y)$ is undirected) and the directed edge $V->Y$ does not belong to the set $\{(Z,Y)$, $(Y,X)$, $(Y,W)$, $X \rightarrow V$, $W \rightarrow V\}$. \\
        For $X$: the directed edge $Z \rightarrow X$ does not belong to the set $\{(Z,Y)$, $(Y,X)$, $(Y,W)$, $X \rightarrow V$, $W \rightarrow V\}$ and the directed edge $V \rightarrow X$ does not belong to the set $\{(Z,Y)$, $(Y,X)$, $(Y,W)$, $X \rightarrow V$, $W \rightarrow V\}$. \\
        For $W$: the directed edge $Z \rightarrow W$ does not belong to the set $\{(Z,Y)$, $(Y,X)$, $(Y,W)$, $X \rightarrow V$, $W \rightarrow V\}$ and the directed edge $V \rightarrow W$ does not belong to the set $\{(Z,Y)$, $(Y,X)$, $(Y,W)$, $X \rightarrow V$, $W \rightarrow V\}$. \\
        From the above, there is no variable in the causal graph $\{(Z,Y)$, $(Y,X)$, $(Y,W)$, $X \rightarrow V$, $W \rightarrow V\}$ with an incoming edge both from $Z$ and another one from $V$, so the hypothesis is False. \\
        \vspace{-1.4mm}
        \textbf{Answer}: 0 \\
        \hline \\
        \vspace{-3mm}
        \underline{\textbf{\textsc{Paraphrased}}} \\
        \vspace{-0.5mm}
        \textbf{Question}: Given the causal graph: $(A,B), (B,C), (B,D), C \rightarrow E, D \rightarrow E$.
        Can you infer if the Hypothesis "There exists at least one collider (i.e., common effect) of $A$ and $B$." is True or False? \\
        \vspace{-1.4mm}
        \textbf{Reasoning}: We have There exists at least one collider (i.e., common effect) of $A$ and $E$.
        We go through all variables apart from $A$ and $E$: $B, C, D$ and check if there is a directed edge from $A$ to it and another one from $E$ to it: \\
        For $B$: the directed edge $A \rightarrow B$ does not belong to the set $\{(A,B)$, $(B,C)$, $(B,D)$, $C \rightarrow E$, $D \rightarrow E\}$ (since $A \rightarrow B$ is directed while $(A,B)$ is undirected) and the directed edge $E \rightarrow B$ does not belong to the set $\{(A,B)$, $(B,C)$, $(B,D)$, $C \rightarrow E$, $D \rightarrow E\}$. \\
        For $C$: the directed edge $A \rightarrow C$ does not belong to the set $\{(A,B)$, $(B,C)$, $(B,D)$, $C \rightarrow E$, $D \rightarrow E\}$ and the directed edge $E \rightarrow C$ does not belong to the set $\{(A,B)$, $(B,C)$, $(B,D)$, $C \rightarrow E$, $D \rightarrow E\}$. \\
        For $D$: the directed edge $A \rightarrow D$ does not belong to the set $\{(A,B)$, $(B,C)$, $(B,D)$, $C \rightarrow E$, $D \rightarrow E\}$ and the directed edge $E \rightarrow D$ does not belong to the set $\{(A,B)$, $(B,C)$, $(B,D)$, $C \rightarrow E$, $D \rightarrow E\}$. \\
        From the above, there is no variable in the causal graph $\{(A,B)$, $(B,C)$, $(B,D)$, $C \rightarrow E$, $D \rightarrow E\}$ with an incoming edge both from $A$ and another one from $E$, so the hypothesis is False. \\
        \vspace{-1.4mm}
        \textbf{Answer}: 0 \\
        \bottomrule
    \end{tabular}
    \label{tab:appendix-pertubing}
\end{table}

\clearpage

\section{\ourprompting on Causal Queries with Natural Stories}\label{sec:appendix_natural}

Tables \ref{tab:appendix-natural-1} and \ref{tab:appendix-natural-2} provide \ourprompting{'s} full output, i.e., reasoning traces and answers for all subquestions, for two natural story scenarios \citep{jin2024can}. The output is correct even though natural story examples were never presented as few-shots. For more details, see the robustness analysis in Sec. \ref{sec:results}.
\begin{table}[h!]
    \caption{    
    \ourprompting reasoning traces and final output, using PaLM 2 L, for the 2-variable natural story example provided in the Appendix of \citet{jin2024can}.
    }
    \centering
    \small
    \begin{tabular}{p{\linewidth}}
        \toprule
        \underline{\textbf{\textsc{Causal Question}}} \\
        \vspace{-1.2mm}
        \textit{Premise}: Suppose there is a closed system of 2 variables, ice cream sales and swimming pool attendance. All the statistical relations among these 2 variables are as follows: ice cream sales correlate with swimming pool attendance.  \\ 
        \textit{Hypothesis}: Ice cream sales directly affect swimming pool attendance. \\
        \toprule
        \vspace{-1.5mm}
        \underline{\textbf{\textsc{\subq1 Reasoning}}} \\
        Since our variables are ice cream sales and swimming pool attendance $\Rightarrow$ the initial fully connected undirected graph is (ice cream sales, swimming pool attendance).  \\
        \vspace{-1.6mm}
        \underline{\textbf{\textsc{Answer}}}: (ice cream sales, swimming pool attendance) \\
        \hline \\
        \vspace{-3.5mm}
        \underline{\textbf{\textsc{\subq2 Reasoning}}} \\
        1. We start with the given fully connected graph: (ice cream sales, swimming pool attendance) \newline
        2. We then check all conditional independencies and remove edges appropriately. In our case: 
        There are no conditional independencies $\Rightarrow$ no edges are removed $\Rightarrow$ the graph after this step is (ice cream sales, swimming pool attendance) \newline
        3. The final undirected graph is: (ice cream sales, swimming pool attendance) \\
        \vspace{-1.6mm}
        \underline{\textbf{\textsc{Answer}}}: (ice cream sales, swimming pool attendance)
        \\
        \hline \\
        \vspace{-3.5mm}
        \underline{\textbf{\textsc{\subq3 Reasoning}}} \\
        We go through all unordered pairs of edges in the undirected graph (ice cream sales, swimming pool attendance) above and find paths of length 2: \\
        Since there is only one edge, there are no paths of length 2. \\
        \vspace{-1.6mm}
        \underline{\textbf{\textsc{Answer}}}: No paths of length 2 found.
        \\
        \hline \\
        \vspace{-3.5mm}
        \underline{\textbf{\textsc{\subq4 Reasoning}}} \\
        No paths of length 2 found $\Rightarrow$ No possible v-structures found. \\
        \vspace{-1.6mm}
        \underline{\textbf{\textsc{Answer}}}: No possible v-structures found
        \\
        \hline \\
        \vspace{-3.5mm}
        \underline{\textbf{\textsc{\subq5 Reasoning}}} \\
        We first go through all possible v-structures: 
        No possible v-structures found. \\
        \vspace{-1.6mm}
        \underline{\textbf{\textsc{Answer}}}: No directed edges found
        \\
        \hline \\
        \vspace{-3.5mm}
        \underline{\textbf{\textsc{\subq6 Reasoning}}} \\
        Since there are no directed edges $\Rightarrow$ the final (undirected) graph remains: (ice cream sales, swimming pool attendance). \\
        \vspace{-1.6mm}
        \underline{\textbf{\textsc{Answer}}}: (ice cream sales, swimming pool attendance)
        \\
        \hline \\
        \vspace{-3.5mm}
        \underline{\textbf{\textsc{\subq7 Reasoning}}} \\
        We go through all paths of length 2 in the graph: No paths of length 2 found $\Rightarrow$ No undirected edge to orient.
        So, the final graph is: (ice cream sales, swimming pool attendance). \\
        \vspace{-1.6mm}
        \underline{\textbf{\textsc{Answer}}}: (ice cream sales, swimming pool attendance)
        \\
        \hline \\
        \vspace{-3.5mm}
        \underline{\textbf{\textsc{\subq8 Reasoning}}} \\
        We have Ice cream sales directly affect swimming pool attendance, that is ice cream sales $\rightarrow$ swimming pool attendance. \\
        The directed edge ice cream sales $\rightarrow$ swimming pool attendance does not belong to the set $\{$(ice cream sales, swimming pool attendance)$\}$ (since ice cream sales $\rightarrow$ swimming pool attendance is directed while (ice cream sales, swimming pool attendance) is undirected), so the hypothesis is False. \\
        \vspace{-1.6mm}
        \underline{\textbf{\textsc{(Final) Answer}}}: 0 \\
        \bottomrule
    \end{tabular}
    \label{tab:appendix-natural-1}
\end{table}

\begin{table}[h!]
    \caption{    
    \ourprompting reasoning traces and final output, using PaLM 2 L, for the 3-variable natural story example provided in the Appendix of \citet{jin2024can}.
    }
    \centering
    \small
    \begin{tabular}{p{\linewidth}}
        \toprule
        \underline{\textbf{\textsc{Causal Question}}} \\
        \vspace{-1.5mm}
        \textit{Premise}: Let’s consider three factors: eating junk food, obesity, and watching television. There is a correlation between eating junk food and obesity, and between watching television and obesity. However, eating junk food and watching television are independent from each other. \\ 
        \textit{Hypothesis}: Eating junk food directly affects obesity. \\
        \toprule
        \vspace{-2mm}
        \underline{\textbf{\textsc{\subq1 Reasoning}}} \\
        Since our variables are eating junk food, obesity, watching television $\Rightarrow$ the initial fully connected undirected graph is (eating junk food, obesity), (eating junk food, watching television), (obesity, watching television).  \\
        \vspace{-1.6mm}
        \underline{\textbf{\textsc{Answer}}}: (eating junk food, obesity), (eating junk food, watching television), (obesity, watching television)\\
        \hline \\
        \vspace{-4mm}
        \underline{\textbf{\textsc{\subq2 Reasoning}}} \\
        \vspace{-3mm}
        1. We start with the given fully connected graph: (eating junk food, obesity), (eating junk food, watching television), (obesity, watching television). \\
        2. We then check all conditional independencies and remove edges appropriately. In our case: 
        eating junk food is independent of watching television $\Rightarrow$ (eating junk food, watching television) is removed $\Rightarrow$ the graph after this step is (eating junk food, obesity), (obesity, watching television). \\
        3. The final undirected graph is: (eating junk food, obesity), (obesity, watching television). \\
        \vspace{-1.6mm}
        \underline{\textbf{\textsc{Answer}}}: (eating junk food, obesity), (obesity, watching television) \\
        \hline \\
        \vspace{-4mm}
        \underline{\textbf{\textsc{\subq3 Reasoning}}} \\
        \vspace{-3mm}
        We go through all unordered pairs of edges in the undirected graph (eating junk food, obesity), (obesity, watching television) above and find paths of length 2: \\
        Since the edges (eating junk food, obesity) and (obesity, watching television) share a common node obesity $\Rightarrow$ (eating junk food, obesity, watching television) is a path of length 2. \\
        \vspace{-1.6mm}
        \underline{\textbf{\textsc{Answer}}}: (eating junk food, obesity, watching television) \\
        \hline \\
        \vspace{-4mm}
        \underline{\textbf{\textsc{\subq4 Reasoning}}} \\
        \vspace{-3mm}
        We go through all paths of length 2 and check if there is an edge connecting the start and end of the path in the undirected graph: (eating junk food, obesity), (obesity, watching television). \\
        For path (eating junk food, obesity, watching television): (eating junk food, watching television) does not belong to the set of edges $\{$(eating junk food, obesity), (obesity, watching television)$\}$ $\Rightarrow$ (eating junk food, obesity, watching television) is a possible v-structure. \\
        \vspace{-1.6mm}
        \underline{\textbf{\textsc{Answer}}}: (eating junk food, obesity, watching television) \\
        \hline \\
        \vspace{-4mm}
        \underline{\textbf{\textsc{\subq5 Reasoning}}} \\
        \vspace{-3mm}
        We first go through all possible v-structures: \\
        \vspace{-3mm}
        (eating junk food, obesity, watching television) is a possible v-structure: all conditional independences between eating junk food and watching television are: eating junk food and watching television are independent from each other. Since the middle variable obesity is not in any conditioning set that makes eating junk food and watching television independent $\Rightarrow$ (eating junk food, obesity, watching television) form a v-structure $\Rightarrow$ we orient the arrows towards the middle node obesity: eating junk food $\rightarrow$ obesity, watching television $\rightarrow$ obesity.  \\
        \vspace{-1.6mm}
        \underline{\textbf{\textsc{Answer}}}: eating junk food $\rightarrow$ obesity, watching television $\rightarrow$ obesity \\
        \hline \\
        \vspace{-4mm}
        \underline{\textbf{\textsc{\subq6 Reasoning}}} \\
        \vspace{-3mm}
        (eating junk food, obesity) becomes eating junk food $\rightarrow$ obesity \newline
        (obesity, watching television) becomes watching television $\rightarrow$ obesity \newline
        So the final (partially) directed graph is: eating junk food $\rightarrow$ obesity, watching television $\rightarrow$ obesity. \\
        \vspace{-1.6mm}
        \underline{\textbf{\textsc{Answer}}}: eating junk food $\rightarrow$ obesity, watching television $\rightarrow$ obesity \\
        \hline \\
        \vspace{-4mm}
        \underline{\textbf{\textsc{\subq7 Reasoning}}} \\
        \vspace{-3mm}
        We go through all paths of length 2 in the graph: (eating junk food, obesity, watching television) and possibly orient non directed edges. In our case: \\
        For path (eating junk food, obesity, watching television) with middle node obesity: Both eating junk food $\rightarrow$ obesity and watching television $\rightarrow$ obesity are directed $\Rightarrow$ No undirected edge to orient. \\
        So, the final graph is: eating junk food $\rightarrow$ obesity, watching television $\rightarrow$ obesity.  \\
        \vspace{-1.6mm}
        \underline{\textbf{\textsc{Answer}}}: eating junk food $\rightarrow$ obesity, watching television $\rightarrow$ obesity
        \\
        \hline \\
        \vspace{-4mm}
        \underline{\textbf{\textsc{\subq8 Reasoning}}} \\
        \vspace{-3mm}
        We have Eating junk food directly affects obesity, that is eating junk food $\rightarrow$ obesity. \\
        The directed edge eating junk food $\rightarrow$ obesity belongs to the set $\{$eating junk food $\rightarrow$ obesity, watching television $\rightarrow$ obesity$\}$, so the hypothesis is True. \\
        \vspace{-1.6mm}
        \underline{\textbf{\textsc{(Final) Answer}}}: 1 \\
        \bottomrule
    \end{tabular}
    \label{tab:appendix-natural-2}
\end{table}

\clearpage
\section{\ourprompting: Transparent and Interpretable  Output Eases Error Tracing}\label{sec:appendix_error_tracing}

By guiding the \llm to reason through applying the steps of the \pc algorithm, we gain more transparent and interpretable results: for instance, it is possible to trace which steps are responsible for an erroneous final answer. Table \ref{tab:appendix-trace-error-a} shows \ourprompting output for the causal query provided in Table \ref{tab:appendix-trace-error-q}. By inspecting the reasonings and answers, we can determine that the erroneous final answer is only due to the final \subq8 (see red highlighted text). Instead, the inferred causal graph after \subq7 is correct (see green highlighted text) as are all previous reasoning traces and answers.

\begin{table}[h!]
    \caption{   
    A 5-variable \ctoc example,
    its \ourprompting output is provided in Table \ref{tab:appendix-trace-error-a}.
    }
    \centering
    \scriptsize
    \begin{tabular}{p{\linewidth}}
        \toprule
        \underline{\textbf{\textsc{Causal Question}}} \\
        \vspace{-1.5mm}
        \textit{Premise}: Suppose there is a closed system of 5 variables, A, B, C, D and E. All the statistical relations among these 5 variables are as follows: A correlates with B. A correlates with C. A correlates with D. A correlates with E. B correlates with C. B correlates with D. B correlates with E. C correlates with D. C correlates with E. D correlates with E. However, A and C are independent given B. A and C are independent given B and D. A and C are independent given B, D and E. A and C are independent given B and E. A and D are independent given B. A and D are independent given B and C. A and D are independent given B, C and E. A and D are independent given B and E. A and D are independent given C. A and D are independent given C and E. B and D are independent given A and C. B and D are independent given A, C and E. B and D are independent given C. B and D are independent given C and E. B and E are independent given A. B and E are independent given A and C. B and E are independent given A, C and D. B and E are independent given A and D. C and E are independent given A. C and E are independent given A and B. C and E are independent given A, B and D. C and E are independent given A and D. C and E are independent given B. C and E are independent given B and D. D and E are independent given A. D and E are independent given A and B. D and E are independent given A, B and C. D and E are independent given A and C. D and E are independent given B. D and E are independent given B and C. D and E are independent given C. \\ 
        \textit{Hypothesis}: B directly affects C. \\
        \bottomrule
    \end{tabular}
    \label{tab:appendix-trace-error-q}
\end{table}

\begin{table}[h!]
    \caption{    
    \ourprompting reasoning traces and final output, using PaLM 2 L, for the 5-variable \ctoc example of Table \ref{tab:appendix-trace-error-q}.
    }
    \centering
    \scriptsize
    \begin{tabular}{p{\linewidth}}
        \toprule
        \vspace{-2mm}
        \underline{\textbf{\textsc{\subq1 Reasoning}}} \\
        Since our variables are A,B,C,D,E => the initial fully connected undirected graph is (A,B), (A,C), (A,D), (A,E), (B,C), (B,D), (B,E), (C,D), (C,E), (D,E). \\
        \vspace{-1.6mm}
        \underline{\textbf{\textsc{Answer}}}: 
        (A,B), (A,C), (A,D), (A,E), (B,C), (B,D), (B,E), (C,D), (C,E), (D,E) \\
        \hline \\
        \vspace{-4mm}
        \underline{\textbf{\textsc{\subq2 Reasoning}}} \\
        \vspace{-3mm}
        1. We start with the given fully connected graph: (A,B), (A,C), (A,D), (A,E), (B,C), (B,D), (B,E), (C,D), (C,E), (D,E).
        2. We then check all conditional independencies and remove edges appropriately. In our case:
        A and C are independent given B => (A,C) is removed => the graph after this step is (A,B), (A,D), (A,E), (B,C), (B,D), (B,E), (C,D), (C,E), (D,E).
        A and C are independent given B and D => (A,C) is removed => the graph after this step is (A,B), (A,D), (A,E), (B,C), (B,D), (B,E), (C,D), (C,E), (D,E).
        A and C are independent given B, D and E => (A,C) is removed => the graph after this step is (A,B), (A,D), (A,E), (B,C), (B,D), (B,E), (C,D), (C,E), (D,E).
        A and C are independent given B and E => (A,C) is removed => the graph after this step is (A,B), (A,D), (A,E), (B,C), (B,D), (B,E), (C,D), (C,E), (D,E).
        A and D are independent given B => (A,D) is removed => the graph after this step is (A,B), (A,E), (B,C), (B,D), (B,E), (C,D), (C,E), (D,E).
        A and D are independent given B and C => (A,D) is removed => the graph after this step is (A,B), (A,E), (B,C), (B,D), (B,E), (C,D), (C,E), (D,E).
        A and D are independent given B, C and E => (A,D) is removed => the graph after this step is (A,B), (A,E), (B,C), (B,D), (B,E), (C,D), (C,E), (D,E).
        A and D are independent given B and E => (A,D) is removed => the graph after this step is (A,B), (A,E), (B,C), (B,D), (B,E), (C,D), (C,E), (D,E).
        A and D are independent given C => (A,D) is removed => the graph after this step is (A,B), (A,E), (B,C), (B,D), (B,E), (C,D), (C,E), (D,E).
        A and D are independent given C and E => (A,D) is removed => the graph after this step is (A,B), (A,E), (B,C), (B,D), (B,E), (C,D), (C,E), (D,E).
        B and D are independent given A and C => (B,D) is removed => the graph after this step is (A,B), (A,E), (B,C), (B,E), (C,D), (C,E), (D,E).
        B and D are independent given A, C and E => (B,D) is removed => the graph after this step is (A,B), (A,E), (B,C), (B,E), (C,D), (C,E), (D,E).
        'B and D are independent given C => (B,D) is removed => the graph after this step is (A,B), (A,E), (B,C), (B,E), (C,D), (C,E), (D,E).
        B and D are independent given C and E => (B,D) is removed => the graph after this step is (A,B), (A,E), (B,C), (B,E), (C,D), (C,E), (D,E).
        B and E are independent given A => (B,E) is removed => the graph after this step is (A,B), (A,E), (B,C), (C,D), (C,E), (D,E).
        B and E are independent given A and C => (B,E) is removed => the graph after this step is (A,B), (A,E), (B,C), (C,D), (C,E), (D,E).
        B and E are independent given A, C and D => (B,E) is removed => the graph after this step is (A,B), (A,E), (B,C), (C,D), (C,E), (D,E).
        B and E are independent given A and D => (B,E) is removed => the graph after this step is (A,B), (A,E), (B,C), (C,D), (C,E), (D,E).
        C and E are independent given A => (C,E) is removed => the graph after this step is (A,B), (A,E), (B,C), (C,D), (D,E).
        C and E are independent given A and B => (C,E) is removed => the graph after this step is (A,B), (A,E), (B,C), (C,D), (D,E).
        C and E are independent given A, B and D => (C,E) is removed => the graph after this step is (A,B), (A,E), (B,C), (C,D), (D,E).
        C and E are independent given A and D => (C,E) is removed => the graph after this step is (A,B), (A,E), (B,C), (C,D), (D,E).
        C and E are independent given B => (C,E) is removed => the graph after this step is (A,B), (A,E), (B,C), (C,D), (D,E).
        C and E are independent given B and D => (C,E) is removed => the graph after this step is (A,B), (A,E), (B,C), (C,D), (D,E).
        D and E are independent given A => (D,E) is removed => the graph after this step is (A,B), (A,E), (B,C), (C,D).
        D and E are independent given A and B => (D,E) is removed => the graph after this step is (A,B), (A,E), (B,C), (C,D).
        D and E are independent given A, B and C => (D,E) is removed => the graph after this step is (A,B), (A,E), (B,C), (C,D).
        D and E are independent given A and C => (D,E) is removed => the graph after this step is (A,B), (A,E), (B,C), (C,D).
        D and E are independent given B => (D,E) is removed => the graph after this step is (A,B), (A,E), (B,C), (C,D).
        D and E are independent given B and C => (D,E) is removed => the graph after this step is (A,B), (A,E), (B,C), (C,D).
        D and E are independent given C => (D,E) is removed => the graph after this step is (A,B), (A,E), (B,C), (C,D).
        3. The final undirected graph is: (A,B), (A,E), (B,C), (C,D).
         \\
        \vspace{-1.6mm}
        \underline{\textbf{\textsc{Answer}}}: 
        (A,B), (A,E), (B,C), (C,D) \\
        \hline \\
        \vspace{-4mm}
        \underline{\textbf{\textsc{\subq3 Reasoning}}} \\
        \vspace{-3mm}
        We go through all unordered pairs of edges in the undirected graph (A,B), (A,E), (B,C), (C,D) above and find paths of length 2:
        Since the edges (A,B) and (A,E) share a common node A => (B,A,E) is a path of length 2.
        Since the edges (A,B) and (B,C) share a common node B => (A,B,C) is a path of length 2.
        Since the edges (A,B) and (C,D) do not have a common node, they do not form a path of length 2.
        Since the edges (A,E) and (B,C) do not have a common node, they do not form a path of length 2.
        Since the edges (A,E) and (C,D) do not have a common node, they do not form a path of length 2.
        Since the edges (B,C) and (C,D) share a common node C => (B,C,D) is a path of length 2. \\
        \vspace{-1.6mm}
        \underline{\textbf{\textsc{Answer}}}: 
        (B,A,E), (A,B,C), (B,C,D) \\
        \hline \\
        \vspace{-4mm}
        \underline{\textbf{\textsc{\subq4 Reasoning}}} \\
        \vspace{-3mm}
        We go through all paths of length 2 and check if there is an edge connecting the start and end of the path in the undirected graph: (A,B), (A,E), (B,C), (C,D).
        For path (B,A,E): (B,E) does not belong to the set of edges {(A,B), (A,E), (B,C), (C,D)} => (B,A,E) is a possible v-structure.
        For path (A,B,C): (A,C) does not belong to the set of edges {(A,B), (A,E), (B,C), (C,D)} => (A,B,C) is a possible v-structure.
        For path (B,C,D): (B,D) does not belong to the set of edges {(A,B), (A,E), (B,C), (C,D)} => (B,C,D) is a possible v-structure. \\
        \vspace{-1.6mm}
        \underline{\textbf{\textsc{Answer}}}: 
        (B,A,E), (A,B,C), (B,C,D) \\
        \hline \\
        \vspace{-4mm}
        \underline{\textbf{\textsc{\subq5 Reasoning}}} \\
        \vspace{-3mm}
        We first go through all possible v-structures:
        (B,A,E) is a possible v-structure: all conditional independences between B and E are: B and E are independent given A. B and E are independent given A and C. B and E are independent given A, C and D. B and E are independent given A and D. Since the middle variable A is in a conditioning set that makes B and E independent => (B,A,E) does not form a v-structure => No directed edges found.
        (A,B,C) is a possible v-structure: all conditional independences between A and C are: A and C are independent given B. A and C are independent given B and D. A and C are independent given B, D and E. A and C are independent given B and E. Since the middle variable B is in a conditioning set that makes A and C independent => (A,B,C) does not form a v-structure => No directed edges found.
        (B,C,D) is a possible v-structure: all conditional independences between B and D are: B and D are independent given A and C. B and D are independent given A, C and E. B and D are independent given C. B and D are independent given C and E. Since the middle variable C is in a conditioning set that makes B and D independent => (B,C,D) does not form a v-structure => No directed edges found. \\
        \vspace{-1.6mm}
        \underline{\textbf{\textsc{Answer}}}: 
        No directed edges found \\
        \hline \\
        \vspace{-4mm}
        \underline{\textbf{\textsc{\subq6 Reasoning}}} \\
        \vspace{-3mm}
        Since there are no directed edges => the final (undirected) graph remains: (A,B), (A,E), (B,C), (C,D). \\
        \vspace{-1.6mm}
        \underline{\textbf{\textsc{Answer}}}: 
        (A,B), (A,E), (B,C), (C,D) \\
        \hline \\
        \vspace{-4mm}
        \underline{\textbf{\textsc{\subq7 Reasoning}}} \\
        \vspace{-3mm}
        We go through all paths of length 2 in the graph: (B,A,E), (A,B,C), (B,C,D) and possibly orient non directed edges. In our case:
        For path (B,A,E) with middle node A: Both (A,B) and (A,E) are undirected => We do not orient an edge.
        For path (A,B,C) with middle node B: Both (A,B) and (B,C) are undirected => We do not orient an edge.
        For path (B,C,D) with middle node C: Both (B,C) and (C,D) are undirected => We do not orient an edge.
        So the final graph is: (A,B), (A,E), (B,C), (C,D).
        \\
        \vspace{-1.6mm}
        \underline{\textbf{\textsc{Answer}}}: 
        {\sethlcolor{lime}\hl{(A,B), (A,E), (B,C), (C,D)}} \\
        \hline \\
        \vspace{-4mm}
        \underline{\textbf{\textsc{\subq8 Reasoning}}} \\
        \vspace{-3mm}
        We have B directly affects C, that is B->C.
        {\sethlcolor{pink}\hl{The directed edge B->C belongs to the set \{(A,B), (A,E), (B,C), (C,D)\}}}, so the hypothesis is True. \\
        \vspace{-1.6mm}
        \underline{\textbf{\textsc{(Final) Answer}}}: 
        1 \\
        \bottomrule
    \end{tabular}
    \label{tab:appendix-trace-error-a}
\end{table}

\end{document}